\documentclass{article}

\usepackage[round, sort]{natbib}
\usepackage{neurips_esque}

\usepackage[utf8]{inputenc}
\usepackage[T1]{fontenc}
\usepackage{hyperref}
\usepackage{url}
\usepackage{booktabs}
\usepackage{amsfonts}
\usepackage{nicefrac}
\usepackage{microtype}
\usepackage{xcolor}

\usepackage[pdftex]{graphicx}
\usepackage{subfigure}

\title{Enhancing Reasoning with Collaboration and Memory}

\author{
  Julie Michelman$^*$ \quad Nasrin Baratalipour$^*$  \quad Matthew Abueg\thanks{Equal contribution} \\
  \\
  Google DeepMind \\
  \\
  \texttt{\{jmichelman,nasrinb,mattea\}@google.com} \\
}

\begin{document}

\maketitle

\begin{abstract}
  We envision a continuous collaborative learning system where groups of LLM agents work together to solve reasoning problems, drawing on memory they collectively build to improve performance as they gain experience. This work establishes the foundations for such a system by studying the interoperability of chain-of-thought reasoning styles, multi-agent collaboration, and memory banks. Extending beyond the identical agents of self-consistency, we introduce varied-context agents with diverse exemplars and a summarizer agent in place of voting. We generate frozen and continuously learned memory banks of exemplars and pair them with fixed, random, and similarity-based retrieval mechanisms. Our systematic study reveals where various methods contribute to reasoning performance of two LLMs on three grounded reasoning tasks, showing that random exemplar selection can often beat more principled approaches, and in some tasks, inclusion of any exemplars serves only to distract both weak and strong models.
\end{abstract}

\section{Introduction}
\label{intro}

A continuous collaborative learning system combines diverse perspectives of multiple LLM agents with memory for constant self-improvement. In-context learning \citep{brown2020language} and chain-of-thought \citep{wei2022chain, kojima2022large} represent significant advances in inference-time techniques for LLM reasoning. Multi-agent methods \citep{wang2023self, liang2023encouraging, suzgun2024meta} encourage the model to consider multiple angles. Augmenting the model with a memory system \citep{paranjape2023art, das2024larimar} allows it to take advantage of previous successes. We bring together these three ideas - chain-of-thought, multi-agent collaboration, and memory, hypothesizing that they will complement each other to enhance reasoning performance.

For reasoning style we consider: direct prompting, zero-shot chain-of-thought \citep{kojima2022large}, few-shot chain-of-thought \citep{wei2022chain}, and analogical prompting \citep{yasunaga2023large}. For multi-agent collaboration we compare: a single greedy agent, self-consistency \citep{wang2023self}, and our \emph{varied-context} agents with diverse few-shot exemplars. We also introduce a \emph{summarizer agent} as alternative to plurality voting. For memory bank training we propose: a frozen set generated by single-agent zero-shot or continuously learned while running multi-agent few-shot. For memory retrieval mechanism we evaluate: a small fixed set, random sampling, and similarity-based retrieval.

Our results show several consistent trends across datasets and ablations, despite some outliers. First, we find that varied-context agents perform similarly to self-consistency. Random retrieval memory with its diversity of exemplars yields higher accuracy than similarity-based retrieval. Frozen memory performs comparably to incrementally-learned memory, while being more efficient to build. Analogical prompting proves more robust to changes in memory and few-shot design choices than standard chain-of-thought. Additionally, distributing exemplars to varied-context agents is more effective than giving them all to a single agent or multiple identical agents. Finally, the summarizer agent is most helpful when the reasoning agents are weaker and less so when they are already strong.

Our contributions are as follows. We: 1) assess the separate and combined effects of reasoning style, multi-agent collaboration, and memory on LLM reasoning performance; 2) propose varied-context agents that go beyond temperature sampling by basing their responses on different exemplars; 3) propose frozen and learned memory systems paired with retrieval mechanisms; 4) incorporate memory into analogical prompting; and 5) introduce a summarizer agent to aggregate multi-agent responses.

\section{Multi-Agent Memory}
\label{our-method}

\begin{figure*}
  \centering
  \includegraphics[width=0.85\linewidth]{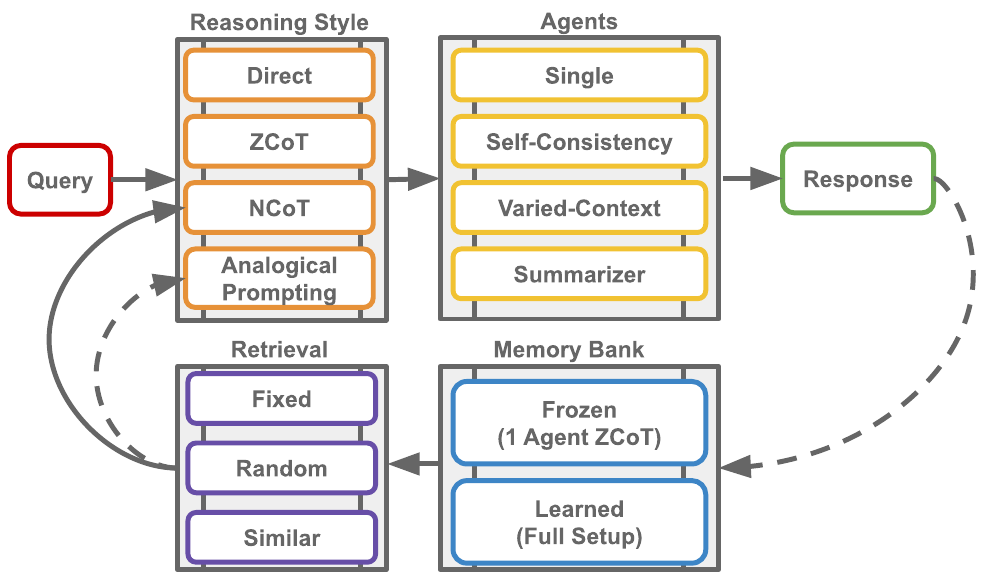}
  \caption{The multi-agent memory reasoning system. Agents each use the given reasoning style: direct, zero- or few-shot chain-of-thought, or analogical prompting. There is a single greedy agent, several temperature-sample agents, or several agents with varied-context (i.e., different exemplars). Answers are combined by voting or a summarizer agent. The memory bank contains exemplars with correct answers from the training set. It can be a reusable frozen set generated by a single greedy ZCoT agent. Or, it can be continuously learned as the multi-agent/in-context learning/retrieval setup utilizes exemplars added to memory earlier in the training pass. Retrieved exemplars are a small fixed set, randomly sampled, or have questions most similar to the current example. Finally, while few-shot CoT requires exemplars, memory is an optional augmentation for analogical prompting.}
  \label{system-diagram}
\end{figure*}

\begin{figure*}
  \centering
  \includegraphics[width=0.95\linewidth]{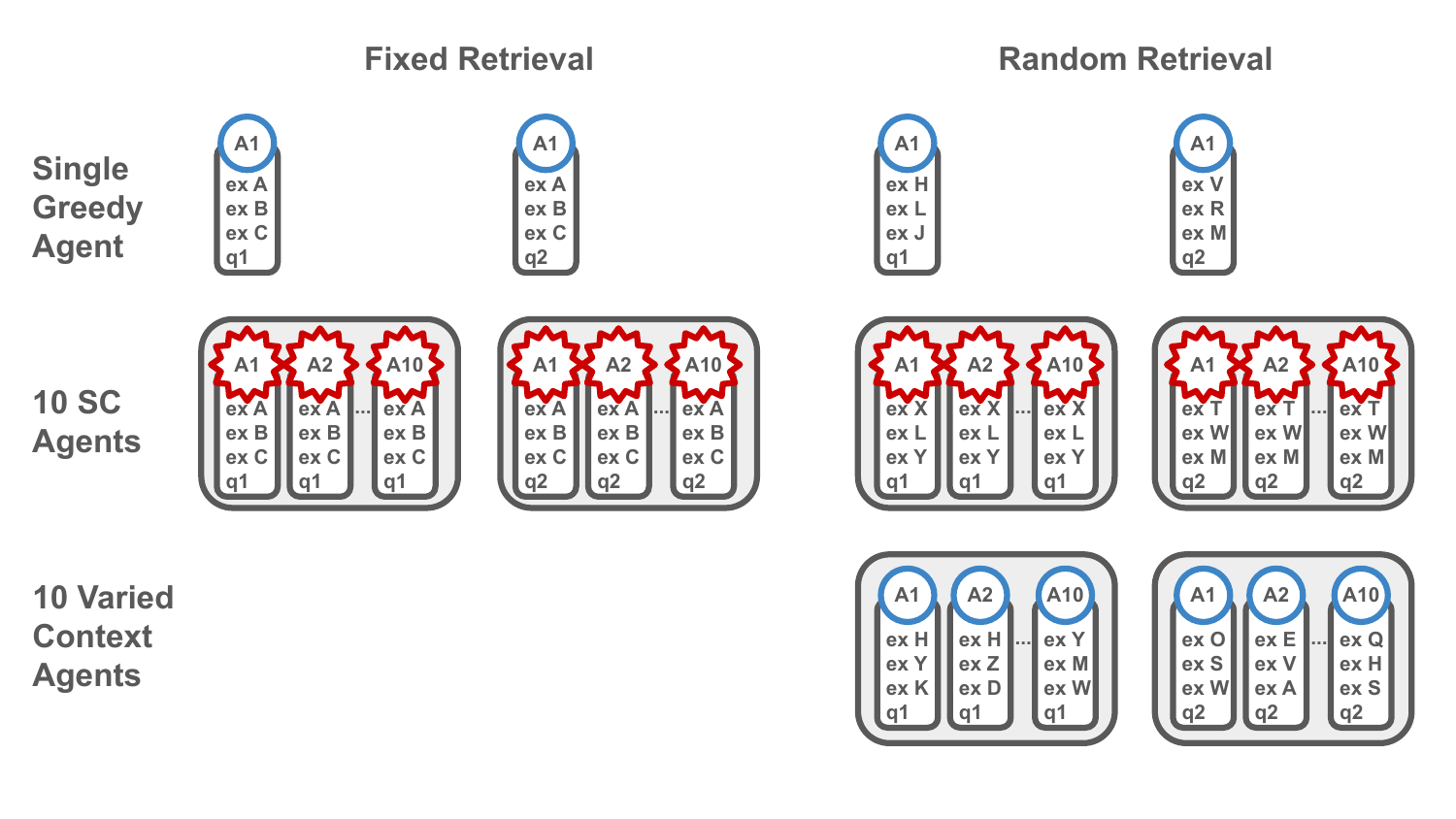}
  \caption{Diagram of agents' NCoT exemplars from a memory bank for two validation set questions. Illustrates three levels of multi-agent collaboration - single, SC, and varied-context - and two types of memory retrieval - fixed and random. A1 is agent 1, ex A is exemplar A, and q1 is question 1. Fixed retrieval reuses the same exemplars every time, while random retrieval samples new exemplars for each question. SC agents share exemplars within a question and use higher temperature (spiky red border) to vary their responses. Varied-context agents independently sample exemplars, so they can use temperature zero (smooth blue border) greedy decoding.}
  \label{agents-diagram}
\end{figure*}

We explore methods to form a holistic reasoning process by varying the style of multi-agent collaboration, chain-of-thought reasoning, and memory, as shown in Figure~\ref{system-diagram}.

\subsection{Multi-Agent Collaboration}
Our baseline is a single agent with \textbf{greedy} decoding. Self-consistency (\textbf{SC}) \citep{wang2023self} is a simple form of multi-agent collaboration: temperature-sampled agents with plurality voting. SC agents are otherwise identical; they all receive the same prompt and in-context learning exemplars. Giving each agent its own perspective can enhance group reasoning by considering the problem from more angles. Perspectives could take the form of roles \citep{zeng2024autodefense}, areas of expertise \citep{suzgun2024meta}, or, as we do here, exemplars in the context. \textbf{Varied-context agents} independently sample exemplars from the memory bank, and again the most common answer is chosen. As an alternative to voting, we introduce a \textbf{summarizer agent} which reviews the others' responses before determining the final answer. It can be paired with identical or varied-context agents.

\subsection{Reasoning Style}
To prompt an agent to solve reasoning problems, we primarily use variants of chain-of-thought. \textbf{Direct} prompting serves as the baseline. Zero-shot chain-of-thought \citep[\textbf{ZCoT},][our Appendix \ref{zcot-prompt}]{kojima2022large} encourages the model to reason on its own, while few-shot chain-of-thought \citep[\textbf{NCoT},][our Appendix \ref{ncot-prompt}]{wei2022chain} places several exemplars in the context. Analogical prompting \citep[\textbf{AP},][our Appendix \ref{analogical-prompt}]{yasunaga2023large} instructs the model to generate its own exemplars then answer the original question within the one LLM call. We extend this idea by inserting exemplars from a memory bank into the AP prompt.

\subsection{Memory Bank}
\label{memory-bank}
We use a memory bank of exemplars for in-context learning \citep{brown2020language}, paired with a retrieval mechanism.

A \textbf{frozen memory bank} is created by running a single greedy ZCoT agent over the training set and keeping examples with correct answers. This mimics a training set with human-written chains-of-thought, but ours are LLM-generated. The smallest memory bank holds a few \textbf{fixed exemplars} that are used every time. To add variety, sample \textbf{random exemplars} from the frozen memory.

A \textbf{learned memory bank} is generated incrementally by running NCoT over the training dataset. For each training example, populate the context with exemplars from the in-progress memory bank. If the model reaches a correct answer, add that question and its chain-of-thought (but not the exemplars it used) to the bank. Figure~\ref{agents-diagram} illustrates several ways exemplars can be assigned to agents. To cold-start the memory bank, begin with ZCoT, then NCoT with one exemplar, and so on, filling as many slots in the NCoT prompt as the memory bank will allow. At validation time no new exemplars are added. Retrieve either \textbf{random exemplars} or the most \textbf{similar exemplars} to the current example - k-nearest neighbors by cosine distance of the questions' Gecko embeddings \citep{lee2024gecko}.

Our AP learned memory bank differs slightly from NCoT. It saves the exemplars the model generates for itself, rather than the correct training example, to more closely mimic the AP system. As with NCoT learned memory, retrieval can be random or by similarity.

\section{Experimental Setup}
\label{experiments}

\subsection{Models}
\label{models}

We run experiments on Gemini 1.0 Pro and Gemini 1.0 Ultra \citep{team2023gemini} models. Self-consistency and identical-context agents preceding the summarizer agent use  temperature sampling (temperature 0.7). The single greedy agent, varied-context agents, and the summarizer agent itself use greedy decoding (temperature 0). The memory embedding system uses the `textembedding-gecko@002' model.

\subsection{Tasks}
\label{tasks}

We use three grounded reasoning benchmarks, where the questions have unambiguous, logical answers. LLM answers are converted to a canonical form before comparing to the label or counting votes for multi-agent collaboration. \textbf{FOLIO} \citep{han2022folio} is a first-order logic task. It presents a series of formal logical statements then asks whether the conclusion is True, False, or Unknown. \textbf{Reasoning About Colored Objects} (RACO) from BIG-bench \citep{srivastava2022beyond} describes a scene involving several objects of various colors, optionally makes a modification, then asks about colors, positions, or counts. \textbf{Tracking Shuffled Objects} (TSO) also from BIG-bench \citep{srivastava2022beyond} assigns several people each an object, performs a series of swaps, then asks which object one of the people has at the end. See Appendix~\ref{task-details} for more details.

\subsection{Combinations of Methods}
\label{combos}

We hypothesize that more sophisticated multi-agent collaboration, reasoning styles, and memory will enhance LLM reasoning. Moreover, we hypothesize that they will complement each other to further improve performance.

\textbf{Main Experiments.} Our main experiments combine multi-agent collaboration (greedy, SC, varied-context agents), reasoning styles (direct, ZCoT, NCoT), and memory (fixed, frozen random, learned random, learned similar exemplars). Not all combinations are possible: direct prompting and ZCoT do not use memory, and varied-context agents must be paired with random retrieval. Our defaults are 10 agents and 3 shots per agent.

\textbf{AP vs CoT.} Compare analogical prompting and chain-of-thought reasoning styles. Pair them with multi-agent collaboration (greedy, SC, varied-context agents) and memory (none, learned random, learned similar exemplars). Compare the original AP against ZCoT, since neither uses exemplars from memory.

\textbf{More Shots vs Varied-Context Agents.} These experiments ablate two aspects of varied-context agents: multiple perspectives and number of exemplars. Consider NCoT with frozen random memory. Holding the number of shots constant at 15 (to avoid context window limits), compare 5 varied-context agents with 3 shots each versus one greedy agent with 15 shots versus 5 SC agents with the same 15 shots as each other.

\textbf{Summarizer Agent vs Voting.} Compare the summarizer agent against multi-agent voting in SC and varied-context. Use a frozen memory bank and the main reasoning styles. Hold constant the number of reasoning agents and thus NCoT exemplars. The summarizer agent is additional and introduces two extra LLM calls - its internal chain-of-thought and final answer.

\section{Results}
\label{results}

\begin{figure*}
  \centering
  \subfigure{\includegraphics[width=0.47\linewidth]{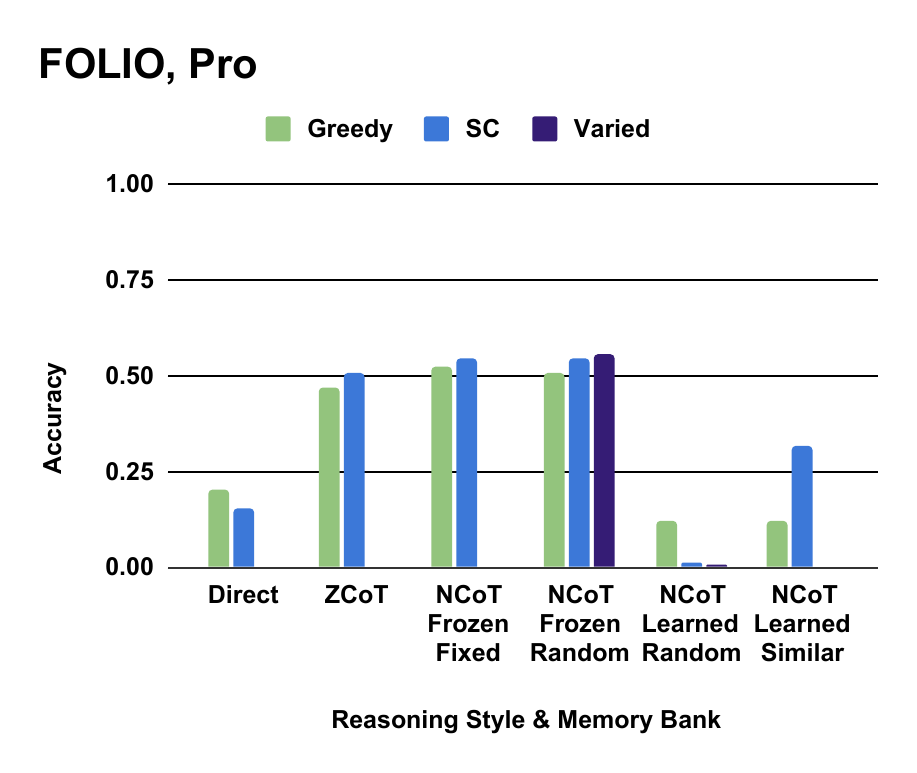}}
  \subfigure{\includegraphics[width=0.47\linewidth]{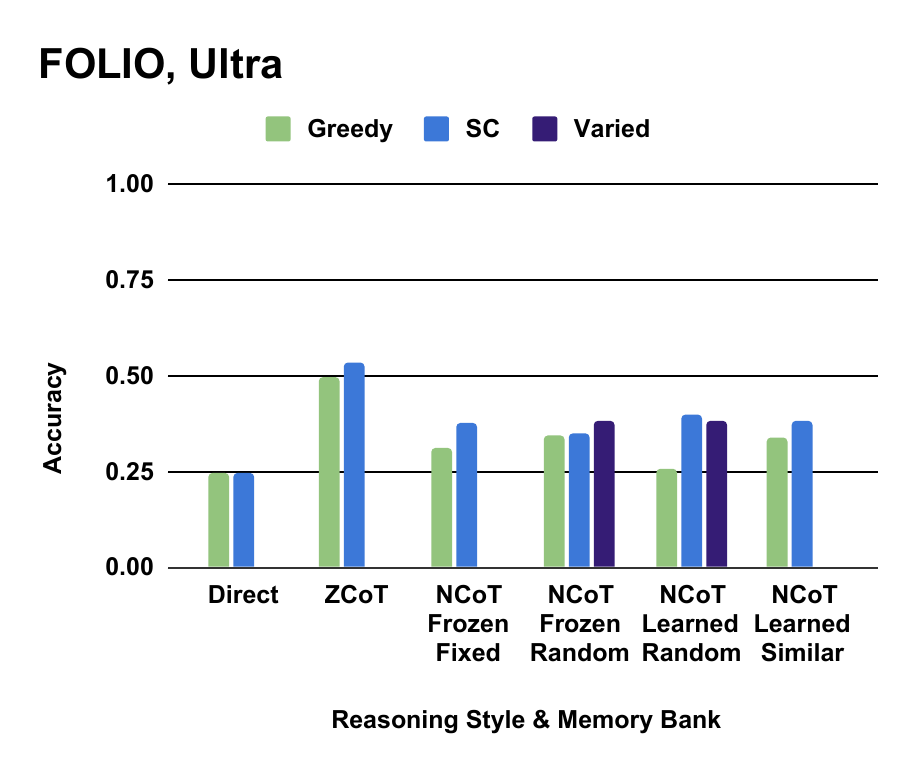}}
  \subfigure{\includegraphics[width=0.47\linewidth]{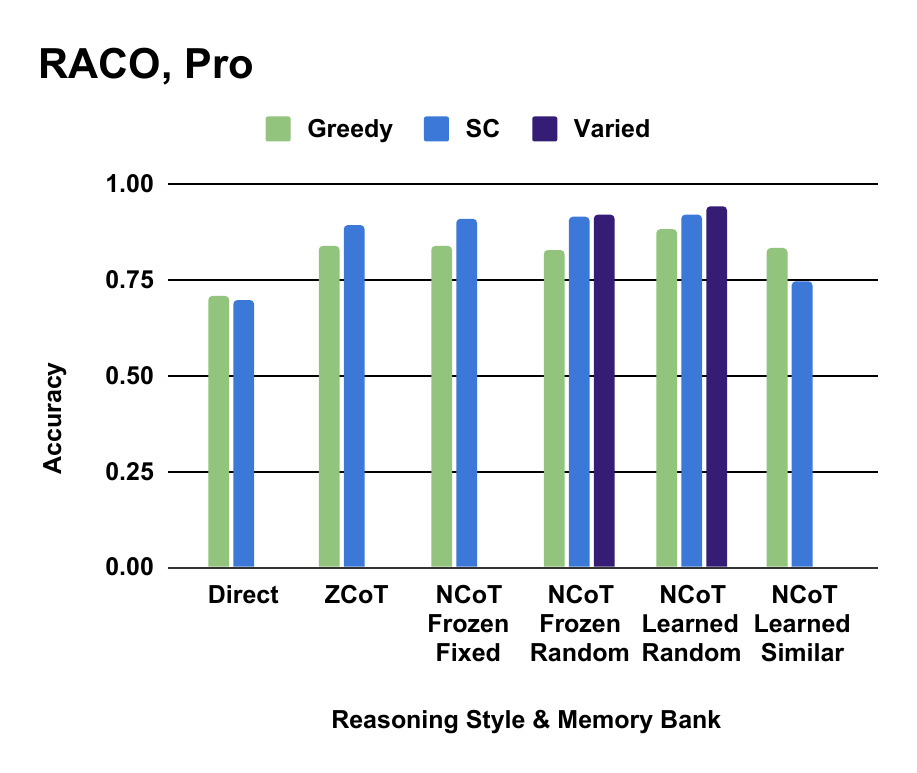}}
  \subfigure{\includegraphics[width=0.47\linewidth]{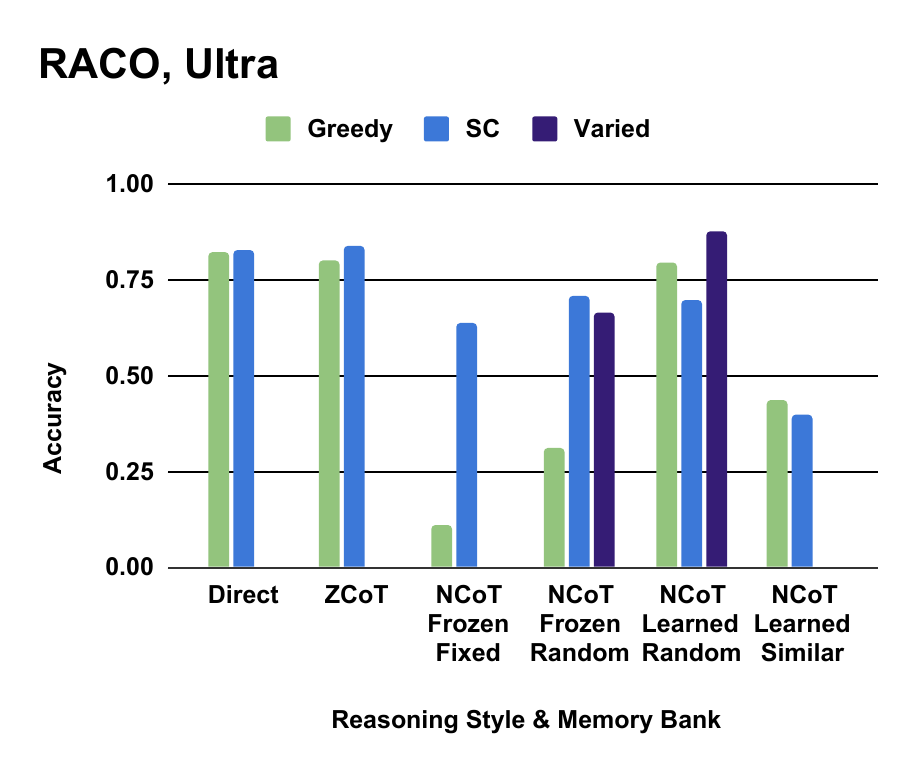}}
  \subfigure{\includegraphics[width=0.47\linewidth]{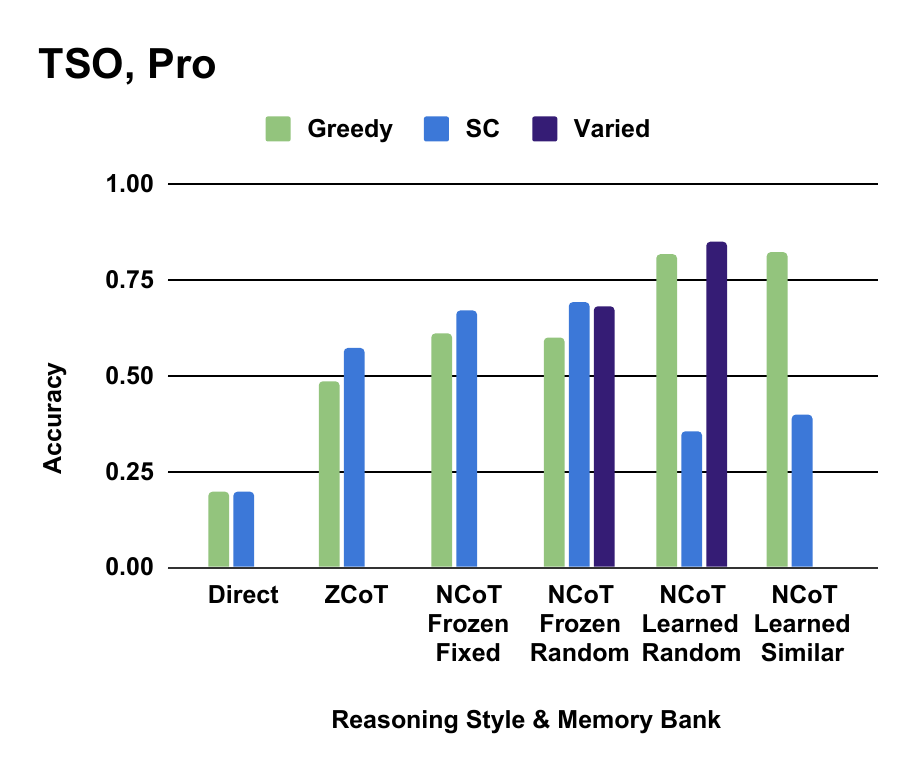}}
  \subfigure{\includegraphics[width=0.47\linewidth]{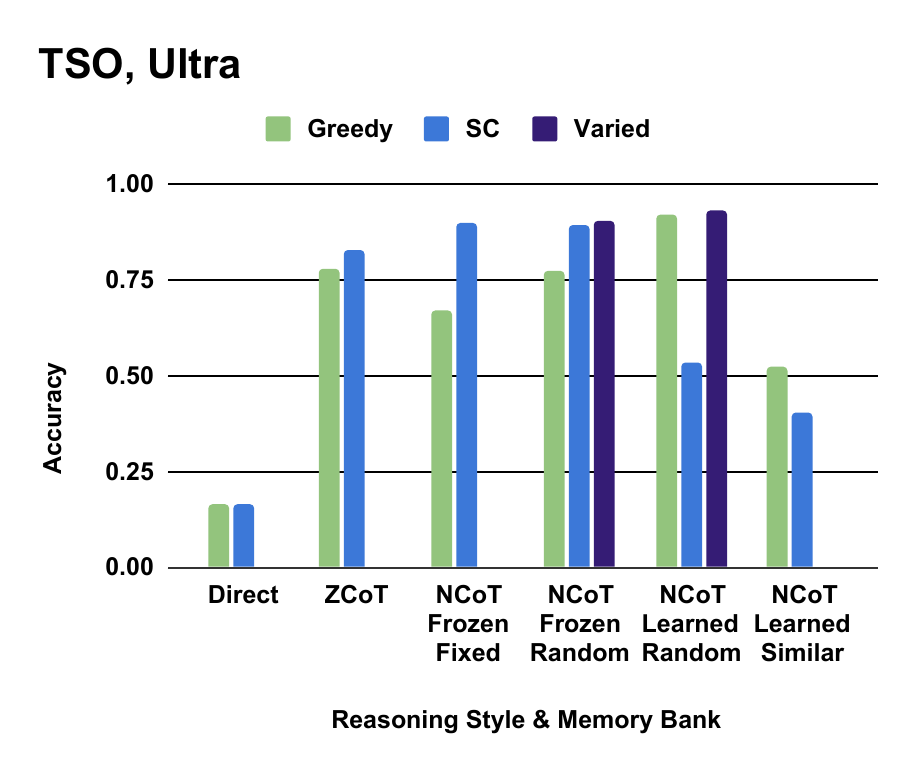}}
  \caption{Main Experiments. Accuracy over the validation set for the two models (Pro and Ultra, in columns) and three tasks (FOLIO, RACO, TSO, in rows). The three concrete reasoning tasks under test are of varying difficulties -- FOLIO - a challenging expert-written, open-domain first-order logic task; RACO - Reasoning about Colored Objects BIG-Bench Hard Q\&A dataset; and TSO - Tracking Shuffled Objects BIG-Bench Hard text completion dataset. Ultra is a significantly larger model than Pro with generally better reasoning performance. ZCoT and NCoT, generally considered standard practice for strong baselines, are beat in many but not all settings by augmented methods.}
  \label{main-plots}
\end{figure*}

\begin{table*}
  \caption{Analogical prompting vs chain-of-thought. Percent accuracy over the validation set, with 2-sigma error bars. Highest accuracy for each task-model pair is displayed in bold. See Figure~\ref{main-plots} and the text for further descriptions of the ablated settings.}
  \label{analogical-table}
  \vskip 0.15in
  \centering
  \begin{tabular}{lllp{0.085\textwidth}p{0.085\textwidth}p{0.085\textwidth}p{0.085\textwidth}p{0.085\textwidth}p{0.085\textwidth}}
    \toprule                                                                                                                                                                                   
                  &                &                 & \multicolumn{6}{c}{\textbf{Reasoning Style and Memory Bank}}                                                                            \\
                                                     \cmidrule(r){4-9}                                                                                                                         
                  &                &                 & ZCoT                  & NCoT                  & NCoT            & AP                    & AP                    & AP                    \\
                  &                &                 &                       & learned               & learned         &                       & learned               & learned               \\
    \textbf{Task} & \textbf{Model} & \textbf{Agents} &                       & random                & similar         &                       & random                & similar               \\
    \midrule                                                                                                                                                                                   
    FOLIO         & Pro            & greedy          & $47\pm\:\:1$          & $12\pm\:\:6$          & $12\pm11$       & $53\pm\:\:1$          & $52\pm\:\:3$          & $50\pm\:\:4$          \\
                  &                & SC              & $51\pm\:\:2$          & $\:\:1\pm\:\:4$       & $32\pm\:\:7$    & $\textbf{56}\pm\:\:2$ & $\textbf{56}\pm\:\:2$ & $\textbf{56}\pm\:\:3$ \\
                  &                & varied          & $-$                   & $\:\:1\pm\:\:1$       & $-$             & $-$                   & $54\pm\:\:1$          & $-$                   \\
                  \cmidrule(r){2-9}                                                                                                                                                            
                  & Ultra          & greedy          & $49\pm\:\:1$          & $26\pm\:\:7$          & $34\pm\:\:5$    & $52\pm\:\:1$          & $50\pm\:\:3$          & $49\pm\:\:3$          \\
                  &                & SC              & $\textbf{54}\pm\:\:1$ & $40\pm\:\:8$          & $38\pm\:\:4$    & $53\pm\:\:1$          & $53\pm\:\:2$          & $52\pm\:\:2$          \\
                  &                & varied          & $-$                   & $38\pm\:\:6$          & $-$             & $-$                   & $52\pm\:\:2$          & $-$                   \\
    \midrule                                                                                                                                                                                   
    RACO          & Pro            & greedy          & $84\pm\:\:0$          & $88\pm\:\:3$          & $83\pm\:\:3$    & $80\pm\:\:1$          & $79\pm\:\:2$          & $82\pm\:\:4$          \\
                  &                & SC              & $89\pm\:\:2$          & $92\pm\:\:2$          & $74\pm10$       & $90\pm\:\:1$          & $90\pm\:\:1$          & $90\pm\:\:2$          \\
                  &                & varied          & $-$                   & $\textbf{94}\pm\:\:2$ & $-$             & $-$                   & $85\pm\:\:2$          & $-$                   \\
                  \cmidrule(r){2-9}                                                                                                                                                            
                  & Ultra          & greedy          & $80\pm\:\:1$          & $79\pm\:\:5$          & $44\pm\:\:4$    & $82\pm\:\:2$          & $82\pm\:\:4$          & $81\pm\:\:2$          \\
                  &                & SC              & $84\pm\:\:2$          & $70\pm24$             & $40\pm\:\:5$    & $83\pm\:\:2$          & $85\pm\:\:2$          & $80\pm\:\:2$          \\
                  &                & varied          & $-$                   & $\textbf{88}\pm\:\:1$ & $-$             & $-$                   & $84\pm\:\:2$          & $-$                   \\
    \midrule                                                                                                                                                                                   
    TSO           & Pro            & greedy          & $49\pm\:\:1$          & $82\pm\:\:2$          & $82\pm\:\:6$    & $36\pm\:\:1$          & $29\pm\:\:9$          & $26\pm\:\:3$          \\
                  &                & SC              & $57\pm\:\:2$          & $36\pm44$             & $40\pm27$       & $46\pm\:\:2$          & $44\pm\:\:2$          & $42\pm\:\:4$          \\
                  &                & varied          & $-$                   & $\textbf{85}\pm\:\:2$ & $-$             & $-$                   & $34\pm\:\:1$          & $-$                   \\
                  \cmidrule(r){2-9}                                                                                                                                                            
                  & Ultra          & greedy          & $78\pm\:\:1$          & $92\pm\:\:1$          & $52\pm\:\:8$    & $25\pm\:\:4$          & $36\pm\:\:4$          & $37\pm17$             \\
                  &                & SC              & $83\pm\:\:1$          & $54\pm97$             & $40\pm61$       & $49\pm\:\:2$          & $59\pm\:\:2$          & $51\pm\:\:8$          \\
                  &                & varied          & $-$                   & $\textbf{93}\pm\:\:1$ & $-$             & $-$                   & $36\pm\:\:3$          & $-$                   \\
    \bottomrule                                                                                                                                                                                
  \end{tabular}
  \vskip -0.1in
\end{table*}

\begin{table*}
  \caption{More shots vs varied-context agents. Percent accuracy over the validation set, with 2-sigma error bars. Highest accuracy for each task-model pair is displayed in bold.}
  \label{allshots-table}
  \vskip 0.15in
  \centering
  \begin{tabular}{p{0.06\textwidth}p{0.06\textwidth}p{0.10\textwidth}p{0.10\textwidth}p{0.10\textwidth}p{0.05\textwidth}p{0.2\textwidth}}
    \toprule                                                                                                                             
                  &                &                 &       &            & \multicolumn{2}{c}{\textbf{Reasoning Style and Memory Bank}} \\
                                                                          \cmidrule(r){6-7}                                              
                  &                & \multicolumn{3}{c}{\textbf{Agents}}  &                   & NCoT                                     \\
                                   \cmidrule(r){3-5}                                                                                     
    \textbf{Task} & \textbf{Model} & type            & count & shots      &                   & frozen random                            \\
    \midrule                                                                                                                             
    FOLIO         & Pro            & greedy          & 1     & 15         &                   & $52\pm\:\:3$                             \\
                  &                & SC              & 5     & 15 shared  &                   & $51\pm\:\:1$                             \\
                  &                & varied          & 5     & 3 each     &                   & $\textbf{56}\pm\:\:2$                    \\
                  \cmidrule(r){2-7}                                                                                                      
                  & Ultra          & greedy          & 1     & 15         &                   & $27\pm\:\:4$                             \\
                  &                & SC              & 5     & 15 shared  &                   & $32\pm\:\:4$                             \\
                  &                & varied          & 5     & 3 each     &                   & $\textbf{36}\pm\:\:4$                    \\
    \midrule                                                                                                                             
    RACO          & Pro            & greedy          & 1     & 15         &                   & $68\pm\:\:4$                             \\
                  &                & SC              & 5     & 15 shared  &                   & $82\pm\:\:4$                             \\
                  &                & varied          & 5     & 3 each     &                   & $\textbf{91}\pm\:\:3$                    \\
                  \cmidrule(r){2-7}                                                                                                      
                  & Ultra          & greedy          & 1     & 15         &                   & $41\pm\:\:6$                             \\
                  &                & SC              & 5     & 15 shared  &                   & $\textbf{62}\pm\:\:3$                    \\
                  &                & varied          & 5     & 3 each     &                   & $51\pm\:\:3$                             \\
    \midrule                                                                                                                             
    TSO           & Pro            & greedy          & 1     & 15         &                   & $57\pm\:\:3$                             \\
                  &                & SC              & 5     & 15 shared  &                   & $59\pm\:\:4$                             \\
                  &                & varied          & 5     & 3 each     &                   & $\textbf{66}\pm\:\:3$                    \\
                  \cmidrule(r){2-7}                                                                                                      
                  & Ultra          & greedy          & 1     & 15         &                   & $46\pm\:\:2$                             \\
                  &                & SC              & 5     & 15 shared  &                   & $64\pm\:\:3$                             \\
                  &                & varied          & 5     & 3 each     &                   & $\textbf{88}\pm\:\:2$                    \\
    \bottomrule                                                                                                                          
  \end{tabular}
  \vskip -0.1in
\end{table*}

\begin{table*}
  \caption{Summarizer agent vs majority voting. Percent accuracy over the validation set, with 2-sigma error bars. Highest accuracy for each task-model pair is displayed in bold. Example diversity provides useful context for the smaller model, but is unnecessary for larger model performance. Summarizing under a simple prompt is more likely to cause distraction than help improve the result for these tasks, except in the case of direct, where it can approximate a crude two-step thought process.}
  \label{summarizer-table}
  \vskip 0.15in
  \centering
  \begin{tabular}{llllp{0.105\textwidth}p{0.105\textwidth}p{0.105\textwidth}p{0.105\textwidth}}
    \toprule                                                                                                                                                             
                  &                &             &        & \multicolumn{4}{c}{\textbf{Reasoning Style and Memory Bank}}                                                 \\
                                                          \cmidrule(r){5-8}                                                                                              
                  &                &             &                       & direct               & ZCoT                  & NCoT                   & NCoT                  \\
                  &                & \multicolumn{2}{c}{\textbf{Agents}} &                      &                       & frozen                 & frozen                \\
                                   \cmidrule(r){3-4}                                                                                                                     
    \textbf{Task} & \textbf{Model} & aggregation & shots                 &                      &                       & fixed                  & random                \\
    \midrule                                                                                                                                                             
    FOLIO         & Pro            & vote        & shared                & $16\pm\:\:2$         & $51\pm\:\:2$          & $54\pm\:\:4$           & $\textbf{55}\pm\:\:3$ \\
                  &                &             & varied                & $-$                  & $-$                   & $-$                    & $\textbf{55}\pm\:\:3$ \\
                  &                & summary     & shared                & $48\pm\:\:3$         & $45\pm\:\:3$          & $41\pm\:\:7$           & $40\pm\:\:6$          \\
                  &                &             & varied                & $-$                  & $-$                   & $-$                    & $44\pm\:\:6$          \\
                  \cmidrule(r){2-8}                                                                                                                                      
                  & Ultra          & vote        & shared                & $25\pm\:\:1$         & $\textbf{54}\pm\:\:1$ & $38\pm\:\:4$           & $35\pm\:\:3$          \\
                  &                &             & varied                & $-$                  & $-$                   & $-$                    & $38\pm\:\:3$          \\
                  &                & summary     & shared                & $34\pm\:\:2$         & $53\pm\:\:3$          & $43\pm\:\:7$           & $46\pm\:\:5$          \\
                  &                &             & varied                & $-$                  & $-$                   & $-$                    & $45\pm\:\:5$          \\
    \midrule                                                                                                                                                             
    RACO          & Pro            & vote        & shared                & $70\pm\:\:1$         & $89\pm\:\:2$          & $91\pm\:\:2$           & $91\pm\:\:1$          \\
                  &                &             & varied                & $-$                  & $-$                   & $-$                    & $\textbf{92}\pm\:\:1$ \\
                  &                & summary     & shared                & $77\pm\:\:2$         & $84\pm\:\:2$          & $69\pm11$              & $69\pm\:\:5$          \\
                  &                &             & varied                & $-$                  & $-$                   & $-$                    & $72\pm\:\:4$          \\
                  \cmidrule(r){2-8}                                                                                                                                      
                  & Ultra          & vote        & shared                & $83\pm\:\:1$         & $\textbf{84}\pm\:\:2$ & $64\pm21$              & $71\pm\:\:4$          \\
                  &                &             & varied                & $-$                  & $-$                   & $-$                    & $66\pm\:\:5$          \\
                  &                & summary     & shared                & $83\pm\:\:2$         & $82\pm\:\:3$          & $79\pm16$              & $76\pm\:\:4$          \\
                  &                &             & varied                & $-$                  & $-$                   & $-$                    & $72\pm\:\:6$          \\
    \midrule                                                                                                                                                             
    TSO           & Pro            & vote        & shared                & $20\pm\:\:2$         & $57\pm\:\:2$          & $67\pm\:\:7$           & $\textbf{69}\pm\:\:2$ \\
                  &                &             & varied                & $-$                  & $-$                   & $-$                    & $68\pm\:\:2$          \\
                  &                & summary     & shared                & $35\pm\:\:2$         & $59\pm\:\:2$          & $42\pm31$              & $41\pm\:\:2$          \\
                  &                &             & varied                & $-$                  & $-$                   & $-$                    & $64\pm\:\:2$          \\
                  \cmidrule(r){2-8}                                                                                                                                      
                  & Ultra          & vote        & shared                & $17\pm\:\:2$         & $83\pm\:\:1$          & $90\pm\:\:4$           & $90\pm\:\:1$          \\
                  &                &             & varied                & $-$                  & $-$                   & $-$                    & $\textbf{91}\pm\:\:1$ \\
                  &                & summary     & shared                & $37\pm\:\:2$         & $82\pm\:\:2$          & $62\pm17$              & $62\pm\:\:3$          \\
                  &                &             & varied                & $-$                  & $-$                   & $-$                    & $65\pm\:\:3$          \\
    \bottomrule                                                                                                                                                          
  \end{tabular}
  \vskip -0.1in
\end{table*}

\subsection{Main Experiments}

The main experiments focus on CoT reasoning styles. NCoT is tested with frozen and learned memory systems. Multi-agent collaboration uses identical and varied-context agents, as well as a single-agent baseline. See Figure~\ref{main-plots} and Table~\ref{main-table} for the full breakdown of primary experiment results.

For reasoning style, ZCoT outperforms direct prompting as expected, but it also beats NCoT in some cases. In terms of multi-agent collaboration, SC has higher accuracy than greedy, except for a few cases with learned memory. Varied-context agents perform similarly to SC agents, rather than offering the boost we expected to see. The effect of memory is less consistent across models and tasks. There is no clear winner among the four variants of memory, but random retrieval tends to outperform similarity-based retrieval. We suspect that the similar exemplars are too similar to each other and the repetitiveness misleads the model. Given the time and complexity of training learned memory, it would be more efficient to use frozen memory.

There are a few case where accuracy drops to near zero. For Ultra on greedy NCoT with fixed exemplars, the exemplars are likely an unlucky sample from the frozen memory. For Pro on multi-agent (SC and varied) NCoT with learned random memory, training accuracy is also low, indicating that the memory bank had difficulty accumulating correct and helpful exemplars.

\subsection{AP vs CoT}
Here we compare analogical prompting to chain-of-thought reasoning - zero-shot and with learned memory. See Table~\ref{analogical-table} and Figure~\ref{analogical-plots}. AP generally outperforms CoT on RACO and especially FOLIO, but CoT is stronger on TSO. AP maintains accuracy level whether it generates its own exemplars on the fly (zero-shot) or retrieves ones created while solving other examples (learned memory). In contrast, ZCoT usually matches or outperforms learned memory NCoT. When the model generates the questions for its exemplars, not just the reasoning, they can have more variety and can be more flexibly applied to other examples.

\subsection{More Shots vs Varied-Context Agents}
We ablate varied-context agents for NCoT reasoning by comparing to greedy and self-consistency with the same total number of exemplars. See Table~\ref{allshots-table} and Figure~\ref{allshots-plots}. For all model-task pairs except Ultra on RACO, varied-context has the highest accuracy of the three, showing that distributing exemplars to multiple agents can prove useful at no more cost than self-consistency. As seen in other experiments, self-consistency has similar or higher accuracy than the single greedy agent. Having the agents different perspectives, whether from temperature sampling or varied exemplars, provides additional value beyond simply including more exemplars in the context.

\subsection{Summarizer Agent vs Voting}
The summarizer agent acts as the aggregator in place of voting in multi-agent collaboration. See Table~\ref{summarizer-table} and Figure~\ref{summarizer-plots}. When the reasoning agents use direct prompting, the summarizer agent increases accuracy beyond voting, as it can approximate a crude two-step thought process. When reasoning agents use ZCoT, it makes little difference, as the summarizer is not provided more useful information. When they use NCoT, voting outperforms the summarizer agent on TSO, as the task becomes further from a text completion, while on FOLIO and RACO it depends on the model, where the weaker model benefits from the additional reasoning. Overall, weaker reasoning agents benefit more from having the summarizer, while voting is enough for stronger reasoning agents.

\section{Related Work}
\label{related-work}

\subsection{Multiple AI Agents}
Recent research has explored leveraging multiple AI agents to augment the capabilities of Large Language Models (LLMs). Studies have demonstrated that multi-agent systems can refine individual LLM outputs, leading to improvements in tasks such as factuality, reasoning, and safety. One basic yet effective approach is ensembling, where multiple responses are generated from an LLM and then integrated. \citet{wang2023self} employ majority voting to determine the final answer. \citet{li2024more} leverage a sampling-and-voting approach to boost the performance of LLMs, particularly in reasoning tasks. Universal Self-Consistency \citep{chen2023universal} uses an LLM agent to extract the majority consensus answer. Our summarizer agent differs in that it performs chain-of-thought reasoning to summarize and determine the best answer. \citet{suzgun2024meta} introduce meta-prompting, a novel scaffolding technique that leverages specialized instances of the same language model, each tailored with specific instructions, to handle the specific subtasks of a large complex task. Debate is another widely used approach, modeling the consideration of diverse ideas. \citet{du2023improving} utilize debate to enhance reasoning and factuality.

\citet{liang2023encouraging} introduced Multi-Agent Debate (MAD), where agents debate a topic, and a judge selects the final solution. Reconcile, proposed by \citet{chen2023reconcile}, aims to improve LLM reasoning abilities by facilitating discussion among multiple LLM agents. It allows them to convince each other to refine their answers through a confidence-weighted voting mechanism to reach a better consensus. \citet{feng2024don} introduce novel approaches based on model collaboration (both cooperative and competitive) to identify knowledge gaps in LLMs, thereby improving their ability to abstain from generating low-confidence outputs. The multi-agent framework has also been widely used in the area of safety. \citet{khan2024debating} employ debate as an effective method for scalable oversight. \citet{zeng2024autodefense} introduced AutoDefense, which uses multiple agents to filter harmful responses and protect LLMs from jailbreak attacks. This is achieved by assigning different roles to LLM agents and employing them collaboratively to complete a defense task.
 
\subsection{Memory}
In-context learning in ML research enables models to adapt based on provided examples, showcasing complex reasoning advancements. But, manual creation of task-specific demonstrations is expensive. \citet{paranjape2023art} introduce a task library with a similarity metric to retrieve relevant examples for LLM inference tasks, aiming to address this issue by building task-specific memories. However, the demonstrations within their library are still hand-crafted. Another form of memory is defined in \citet{das2024larimar} where a distributed episodic memory is defined to augmenting LLMs as a means to online knowledge adaptation. \citet{agarwal2024many} uses LLM-generated CoT exemplars in many-shot Reinforced ICL. Our frozen memory bank is generated in much the same way, but we place a small random sample in the context instead of scaling up to the whole training set.

\section{Limitations and Future Work}
\label{limitations}

This paper evaluates methods on three logical reasoning datasets and uses two LLMs from the same family (Gemini 1.0), so the scope of claims is limited. For example, \citet{bianchi2024well} has demonstrated that behavior in negotiation settings differs across model families. Although we attempt to cover a range of standard reasoning tasks, future work should consider a broader variety of reasoning benchmarks: mathematical, commonsense reasoning, compositional generalization, perhaps even retrieval-based or multi-modal reasoning. Additionally, there is potential headroom in prompt modifications for intermediate model calls.

\section{Conclusion}
\label{conclusion}

We explore the components of a continuous collaborative learning system: reasoning style, memory, and multi-agent collaboration. We introduce varied-context agents that use different exemplars, frozen and learned memory banks, a summarizer agent that aggregate other agents' responses, and a way to incorporate memory into arbitrary exemplar-based methods including analogical prompting. Our experiments show that variety has value. Random retrieval outperforms similarity-based memory, analogical prompting's model-generated exemplars are more robust to reuse, and distributing a fixed number of exemplars among varied-context agents is more effective. Despite some results not matching our initial hypotheses, our methodical analysis sheds light on the strengths and opportunities in multi-agent memory. Our analysis highlights complexities of in-context learning strategies, nuances in strategy selection, and ways in which such strategies can cooperate towards better reasoning performance. It shows that prompt and evaluation construction is critical and provides guidance for potential improvements with increased inference budgets.

\section*{Acknowledgments}
We would like to thank Shawn O'Banion, Abhinav Rastogi, Gil Fidel, Yael Karov, other colleagues at Google DeepMind, and our anonymous reviewers for valuable discussion and feedback on this work.

{\small \bibliography{main}}

\begin{thebibliography}{24}
\providecommand{\natexlab}[1]{#1}
\providecommand{\url}[1]{\texttt{#1}}
\expandafter\ifx\csname urlstyle\endcsname\relax
  \providecommand{\doi}[1]{doi: #1}\else
  \providecommand{\doi}{doi: \begingroup \urlstyle{rm}\Url}\fi

\bibitem[Agarwal et~al.(2024)Agarwal, Singh, Zhang, Bohnet, Chan, Anand, Abbas,
  Nova, Co-Reyes, Chu, et~al.]{agarwal2024many}
Rishabh Agarwal, Avi Singh, Lei~M Zhang, Bernd Bohnet, Stephanie Chan, Ankesh
  Anand, Zaheer Abbas, Azade Nova, John~D Co-Reyes, Eric Chu, et~al.
\newblock Many-shot in-context learning.
\newblock \emph{arXiv preprint arXiv:2404.11018}, 2024.

\bibitem[Bianchi et~al.(2024)Bianchi, Chia, Yuksekgonul, Tagliabue, Jurafsky,
  and Zou]{bianchi2024well}
Federico Bianchi, Patrick~John Chia, Mert Yuksekgonul, Jacopo Tagliabue, Dan
  Jurafsky, and James Zou.
\newblock How well can {LLM}s negotiate? {N}egotiation{A}rena platform and
  analysis.
\newblock \emph{arXiv preprint arXiv:2402.05863}, 2024.

\bibitem[Brown(2020)]{brown2020language}
Tom~B Brown.
\newblock Language models are few-shot learners.
\newblock \emph{arXiv preprint ArXiv:2005.14165}, 2020.

\bibitem[Chen et~al.(2023{\natexlab{a}})Chen, Saha, and
  Bansal]{chen2023reconcile}
Justin Chih-Yao Chen, Swarnadeep Saha, and Mohit Bansal.
\newblock Reconcile: Round-table conference improves reasoning via consensus
  among diverse {LLM}s.
\newblock \emph{arXiv preprint arXiv:2309.13007}, 2023{\natexlab{a}}.

\bibitem[Chen et~al.(2023{\natexlab{b}})Chen, Aksitov, Alon, Ren, Xiao, Yin,
  Prakash, Sutton, Wang, and Zhou]{chen2023universal}
Xinyun Chen, Renat Aksitov, Uri Alon, Jie Ren, Kefan Xiao, Pengcheng Yin,
  Sushant Prakash, Charles Sutton, Xuezhi Wang, and Denny Zhou.
\newblock Universal {S}elf-{C}onsistency for large language model generation.
\newblock \emph{arXiv preprint arXiv:2311.17311}, 2023{\natexlab{b}}.

\bibitem[Das et~al.(2024)Das, Chaudhury, Nelson, Melnyk, Swaminathan, Dai,
  Lozano, Kollias, Chenthamarakshan, Dan, et~al.]{das2024larimar}
Payel Das, Subhajit Chaudhury, Elliot Nelson, Igor Melnyk, Sarath Swaminathan,
  Sihui Dai, Aur{\'e}lie Lozano, Georgios Kollias, Vijil Chenthamarakshan,
  Soham Dan, et~al.
\newblock Larimar: Large language models with episodic memory control.
\newblock \emph{arXiv preprint arXiv:2403.11901}, 2024.

\bibitem[Du et~al.(2023)Du, Li, Torralba, Tenenbaum, and
  Mordatch]{du2023improving}
Yilun Du, Shuang Li, Antonio Torralba, Joshua~B Tenenbaum, and Igor Mordatch.
\newblock Improving factuality and reasoning in language models through
  multiagent debate.
\newblock \emph{arXiv preprint arXiv:2305.14325}, 2023.

\bibitem[Feng et~al.(2024)Feng, Shi, Wang, Ding, Balachandran, and
  Tsvetkov]{feng2024don}
Shangbin Feng, Weijia Shi, Yike Wang, Wenxuan Ding, Vidhisha Balachandran, and
  Yulia Tsvetkov.
\newblock Don't hallucinate, abstain: Identifying {LLM} knowledge gaps via
  multi-{LLM} collaboration.
\newblock \emph{arXiv preprint arXiv:2402.00367}, 2024.

\bibitem[{Gemini Team} et~al.(2023){Gemini Team}, Anil, Borgeaud, Wu, Alayrac,
  Yu, Soricut, Schalkwyk, Dai, Hauth, et~al.]{team2023gemini}
{Gemini Team}, Rohan Anil, Sebastian Borgeaud, Yonghui Wu, Jean-Baptiste
  Alayrac, Jiahui Yu, Radu Soricut, Johan Schalkwyk, Andrew~M Dai, Anja Hauth,
  et~al.
\newblock Gemini: A family of highly capable multimodal models.
\newblock \emph{arXiv preprint arXiv:2312.11805}, 2023.

\bibitem[Han et~al.(2022)Han, Schoelkopf, Zhao, Qi, Riddell, Benson, Sun,
  Zubova, Qiao, Burtell, et~al.]{han2022folio}
Simeng Han, Hailey Schoelkopf, Yilun Zhao, Zhenting Qi, Martin Riddell, Luke
  Benson, Lucy Sun, Ekaterina Zubova, Yujie Qiao, Matthew Burtell, et~al.
\newblock {FOLIO}: Natural language reasoning with first-order logic.
\newblock \emph{arXiv preprint arXiv:2209.00840}, 2022.

\bibitem[Huang et~al.(2023)Huang, Chen, Mishra, Zheng, Yu, Song, and
  Zhou]{huang2023large}
Jie Huang, Xinyun Chen, Swaroop Mishra, Huaixiu~Steven Zheng, Adams~Wei Yu,
  Xinying Song, and Denny Zhou.
\newblock Large language models cannot self-correct reasoning yet.
\newblock \emph{arXiv preprint arXiv:2310.01798}, 2023.

\bibitem[Khan et~al.(2024)Khan, Hughes, Valentine, Ruis, Sachan, Radhakrishnan,
  Grefenstette, Bowman, Rockt{\"a}schel, and Perez]{khan2024debating}
Akbir Khan, John Hughes, Dan Valentine, Laura Ruis, Kshitij Sachan, Ansh
  Radhakrishnan, Edward Grefenstette, Samuel~R Bowman, Tim Rockt{\"a}schel, and
  Ethan Perez.
\newblock Debating with more persuasive {LLM}s leads to more truthful answers.
\newblock \emph{arXiv preprint arXiv:2402.06782}, 2024.

\bibitem[Kojima et~al.(2022)Kojima, Gu, Reid, Matsuo, and
  Iwasawa]{kojima2022large}
Takeshi Kojima, Shixiang~Shane Gu, Machel Reid, Yutaka Matsuo, and Yusuke
  Iwasawa.
\newblock Large language models are zero-shot reasoners.
\newblock \emph{Advances in Neural Information Processing Systems},
  35:\penalty0 22199--22213, 2022.

\bibitem[Lee et~al.(2024)Lee, Dai, Ren, Chen, Cer, Cole, Hui, Boratko, Kapadia,
  Ding, et~al.]{lee2024gecko}
Jinhyuk Lee, Zhuyun Dai, Xiaoqi Ren, Blair Chen, Daniel Cer, Jeremy~R Cole, Kai
  Hui, Michael Boratko, Rajvi Kapadia, Wen Ding, et~al.
\newblock Gecko: Versatile text embeddings distilled from large language
  models.
\newblock \emph{arXiv preprint arXiv:2403.20327}, 2024.
\newblock URL
  \url{https://cloud.google.com/vertex-ai/generative-ai/docs/model-reference/text-embeddings-api#model_versions}.

\bibitem[Li et~al.(2024)Li, Zhang, Yu, Fu, and Ye]{li2024more}
Junyou Li, Qin Zhang, Yangbin Yu, Qiang Fu, and Deheng Ye.
\newblock More agents is all you need.
\newblock \emph{arXiv preprint arXiv:2402.05120}, 2024.

\bibitem[Liang et~al.(2023)Liang, He, Jiao, Wang, Wang, Wang, Yang, Tu, and
  Shi]{liang2023encouraging}
Tian Liang, Zhiwei He, Wenxiang Jiao, Xing Wang, Yan Wang, Rui Wang, Yujiu
  Yang, Zhaopeng Tu, and Shuming Shi.
\newblock Encouraging divergent thinking in large language models through
  multi-agent debate.
\newblock \emph{arXiv preprint arXiv:2305.19118}, 2023.

\bibitem[Paranjape et~al.(2023)Paranjape, Lundberg, Singh, Hajishirzi,
  Zettlemoyer, and Ribeiro]{paranjape2023art}
Bhargavi Paranjape, Scott Lundberg, Sameer Singh, Hannaneh Hajishirzi, Luke
  Zettlemoyer, and Marco~Tulio Ribeiro.
\newblock {ART}: Automatic multi-step reasoning and tool-use for large language
  models.
\newblock \emph{arXiv preprint arXiv:2303.09014}, 2023.

\bibitem[Srivastava et~al.(2022)Srivastava, Rastogi, Rao, Shoeb, Abid, Fisch,
  Brown, Santoro, Gupta, Garriga-Alonso, et~al.]{srivastava2022beyond}
Aarohi Srivastava, Abhinav Rastogi, Abhishek Rao, Abu Awal~Md Shoeb, Abubakar
  Abid, Adam Fisch, Adam~R Brown, Adam Santoro, Aditya Gupta, Adri{\`a}
  Garriga-Alonso, et~al.
\newblock Beyond the imitation game: Quantifying and extrapolating the
  capabilities of language models.
\newblock \emph{arXiv preprint arXiv:2206.04615}, 2022.

\bibitem[Suzgun and Kalai(2024)]{suzgun2024meta}
Mirac Suzgun and Adam~Tauman Kalai.
\newblock Meta-prompting: Enhancing language models with task-agnostic
  scaffolding.
\newblock \emph{arXiv preprint arXiv:2401.12954}, 2024.

\bibitem[Suzgun et~al.(2022)Suzgun, Scales, Sch{\"a}rli, Gehrmann, Tay, Chung,
  Chowdhery, Le, Chi, Zhou, et~al.]{suzgun2022challenging}
Mirac Suzgun, Nathan Scales, Nathanael Sch{\"a}rli, Sebastian Gehrmann, Yi~Tay,
  Hyung~Won Chung, Aakanksha Chowdhery, Quoc~V Le, Ed~H Chi, Denny Zhou, et~al.
\newblock Challenging {BIG}-bench tasks and whether chain-of-thought can solve
  them.
\newblock \emph{arXiv preprint arXiv:2210.09261}, 2022.

\bibitem[Wang et~al.(2023)Wang, Wei, Schuurmans, Le, Chi, Narang, Chowdhery,
  and Zhou]{wang2023self}
X~Wang, J~Wei, D~Schuurmans, Q~Le, E~Chi, S~Narang, A~Chowdhery, and D~Zhou.
\newblock Self-consistency improves chain of thought reasoning in language
  models.
\newblock \emph{arXiv preprint arXiv:2203.11171}, 2023.

\bibitem[Wei et~al.(2022)Wei, Wang, Schuurmans, Bosma, Xia, Chi, Le, Zhou,
  et~al.]{wei2022chain}
Jason Wei, Xuezhi Wang, Dale Schuurmans, Maarten Bosma, Fei Xia, Ed~Chi, Quoc~V
  Le, Denny Zhou, et~al.
\newblock Chain-of-thought prompting elicits reasoning in large language
  models.
\newblock \emph{Advances in Neural Information Processing Systems},
  35:\penalty0 24824--24837, 2022.

\bibitem[Yasunaga et~al.(2023)Yasunaga, Chen, Li, Pasupat, Leskovec, Liang,
  Chi, and Zhou]{yasunaga2023large}
Michihiro Yasunaga, Xinyun Chen, Yujia Li, Panupong Pasupat, Jure Leskovec,
  Percy Liang, Ed~H Chi, and Denny Zhou.
\newblock Large language models as analogical reasoners.
\newblock \emph{arXiv preprint arXiv:2310.01714}, 2023.

\bibitem[Zeng et~al.(2024)Zeng, Wu, Zhang, Wang, and Wu]{zeng2024autodefense}
Yifan Zeng, Yiran Wu, Xiao Zhang, Huazheng Wang, and Qingyun Wu.
\newblock Autodefense: Multi-agent {LLM} defense against jailbreak attacks.
\newblock \emph{arXiv preprint arXiv:2403.04783}, 2024.

\end{thebibliography}

\newpage
\appendix

\section{Additional Tables and Plots}
\label{additional-tables}

\begin{figure*}
  \centering
  \subfigure{\includegraphics[width=0.44\linewidth]{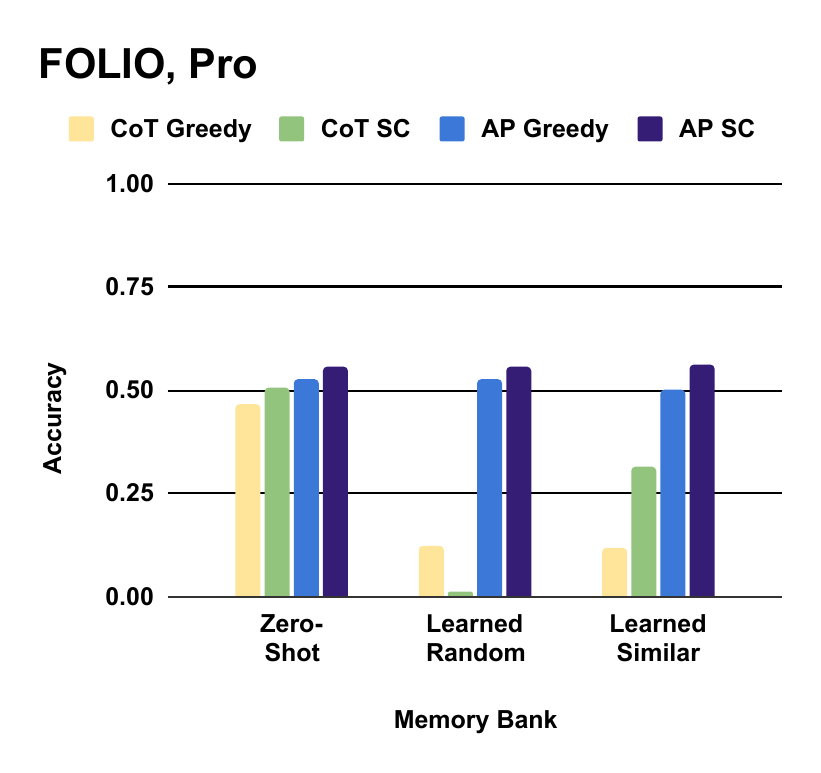}}
  \subfigure{\includegraphics[width=0.44\linewidth]{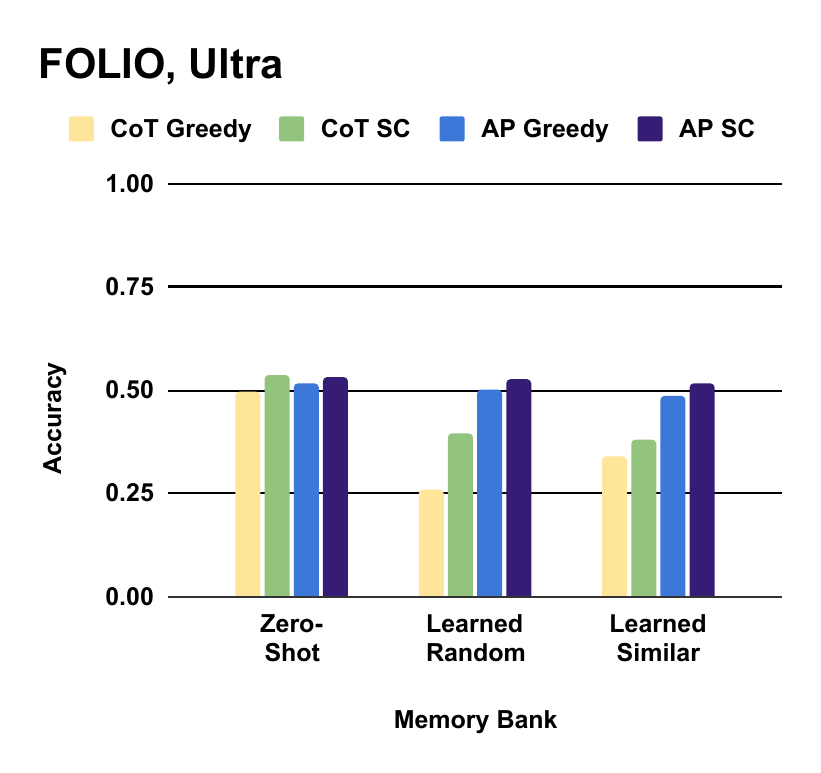}}
  \subfigure{\includegraphics[width=0.44\linewidth]{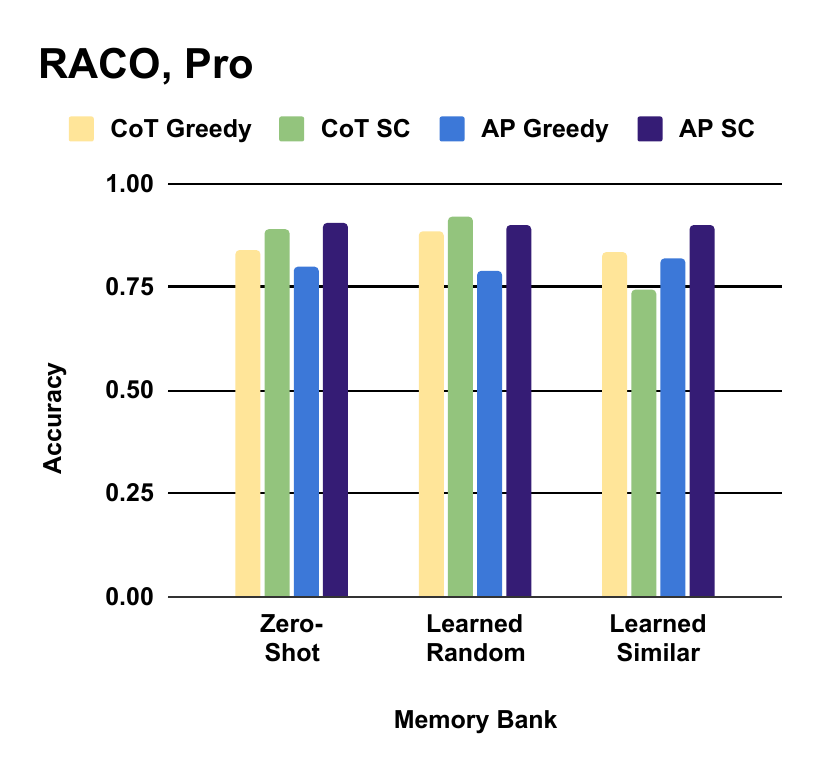}}
  \subfigure{\includegraphics[width=0.44\linewidth]{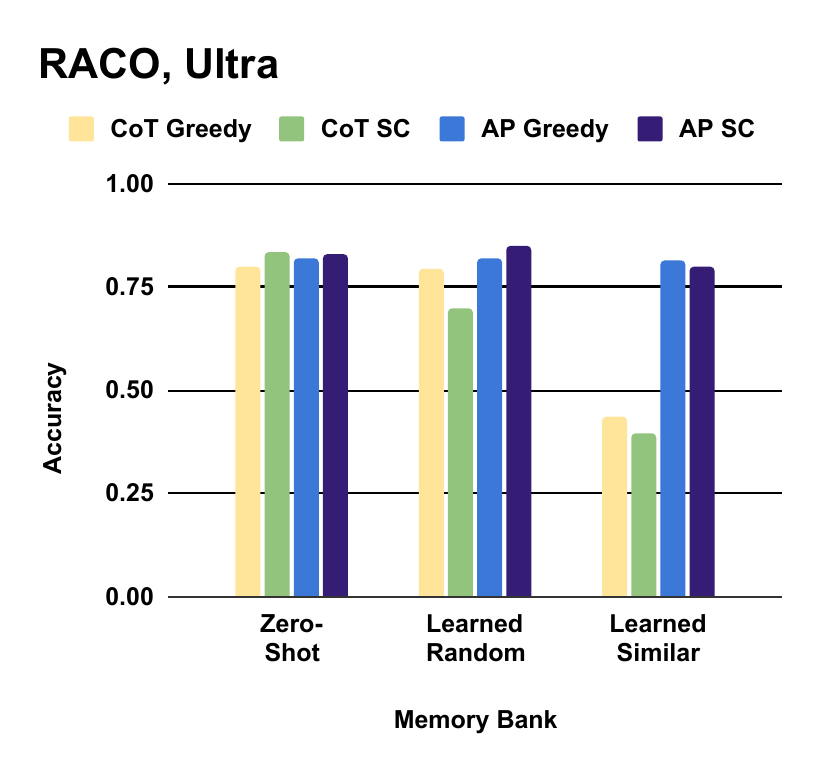}}
  \subfigure{\includegraphics[width=0.44\linewidth]{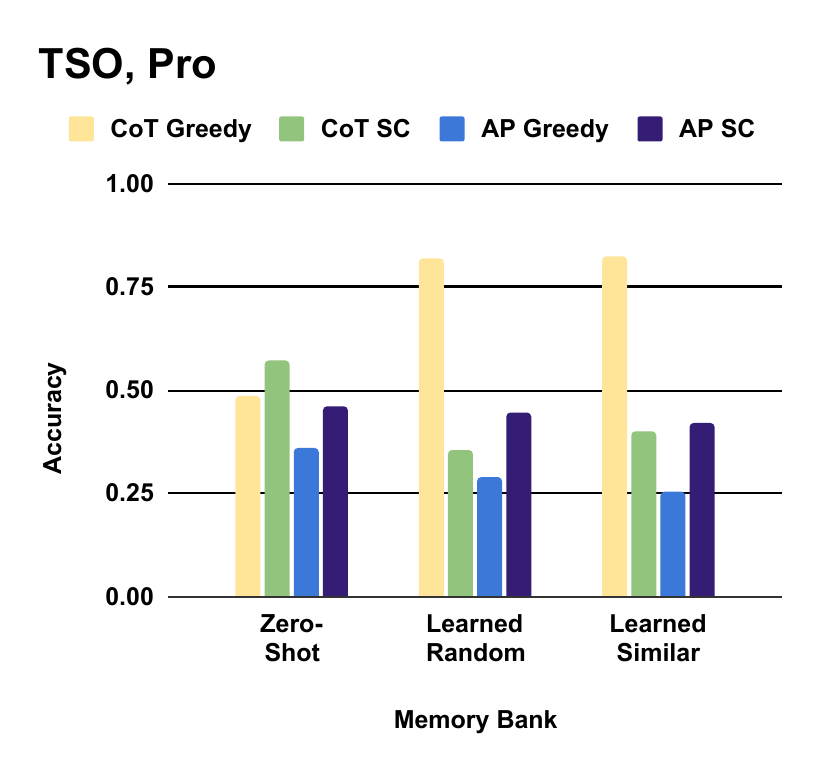}}
  \subfigure{\includegraphics[width=0.44\linewidth]{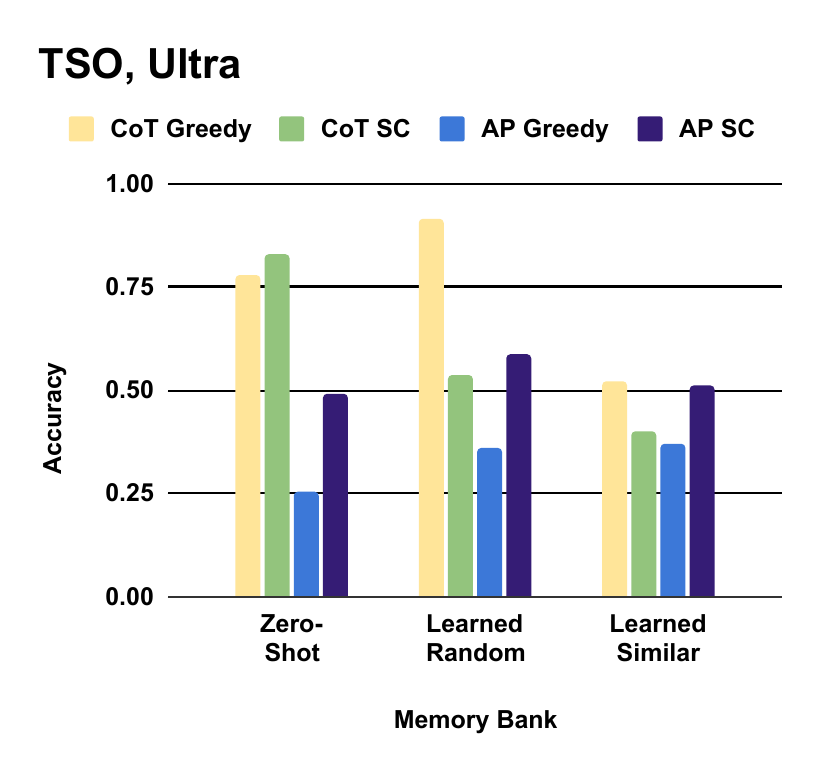}}
  \caption{Analogical Prompting vs Chain-of-Thought. Accuracy over the validation set for the two models and three tasks.}
  \label{analogical-plots}
\end{figure*}

\begin{figure*}
  \centering
  \subfigure{\includegraphics[width=0.49\linewidth]{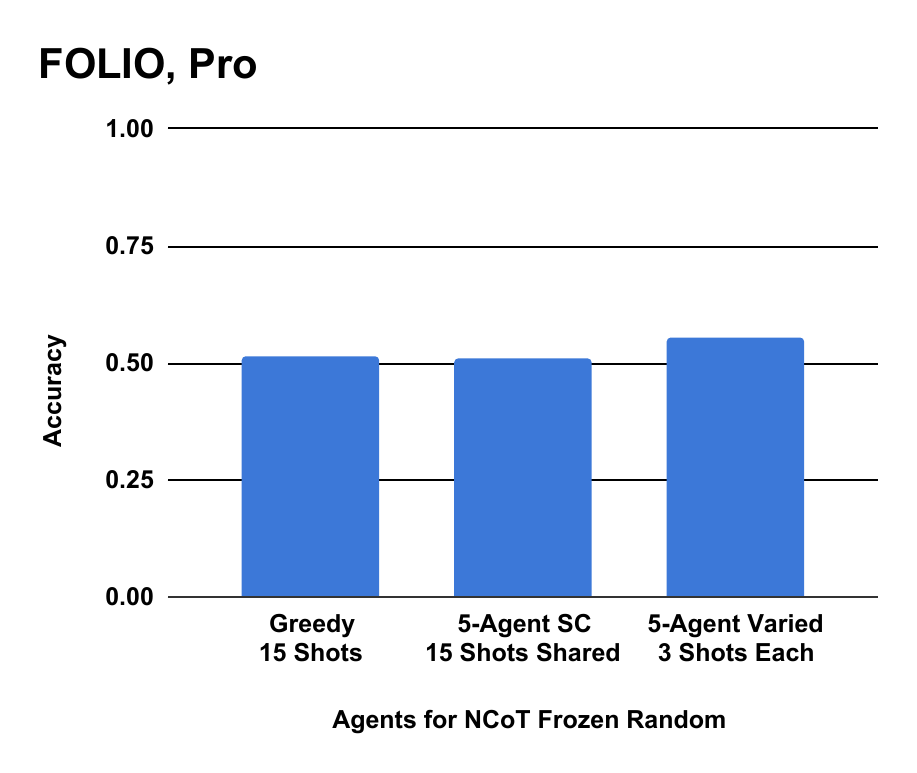}}
  \subfigure{\includegraphics[width=0.49\linewidth]{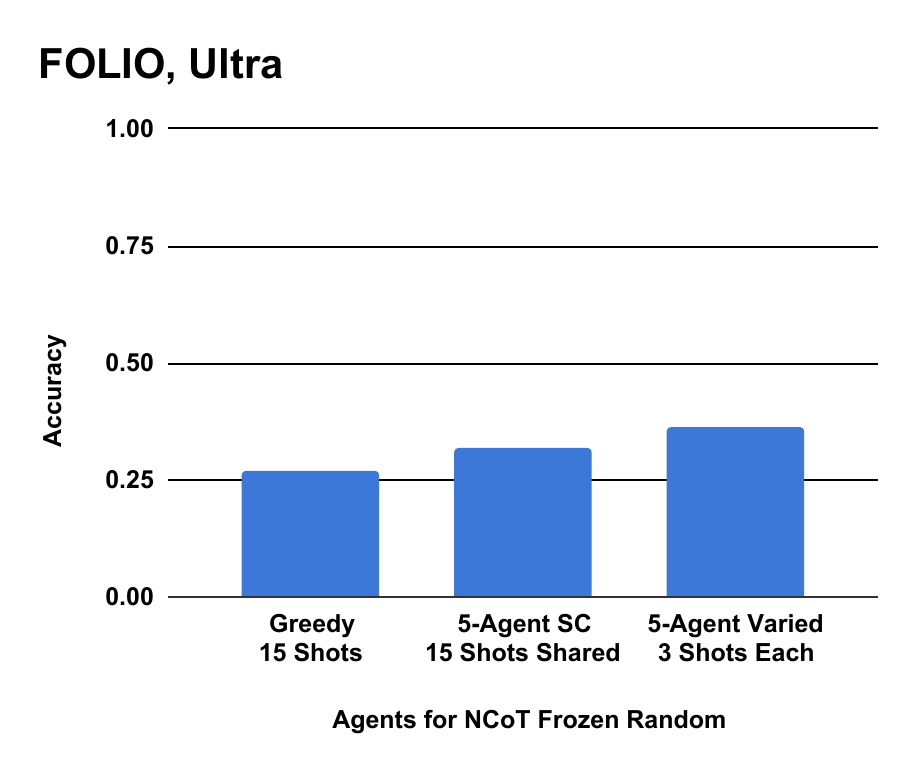}}
  \subfigure{\includegraphics[width=0.49\linewidth]{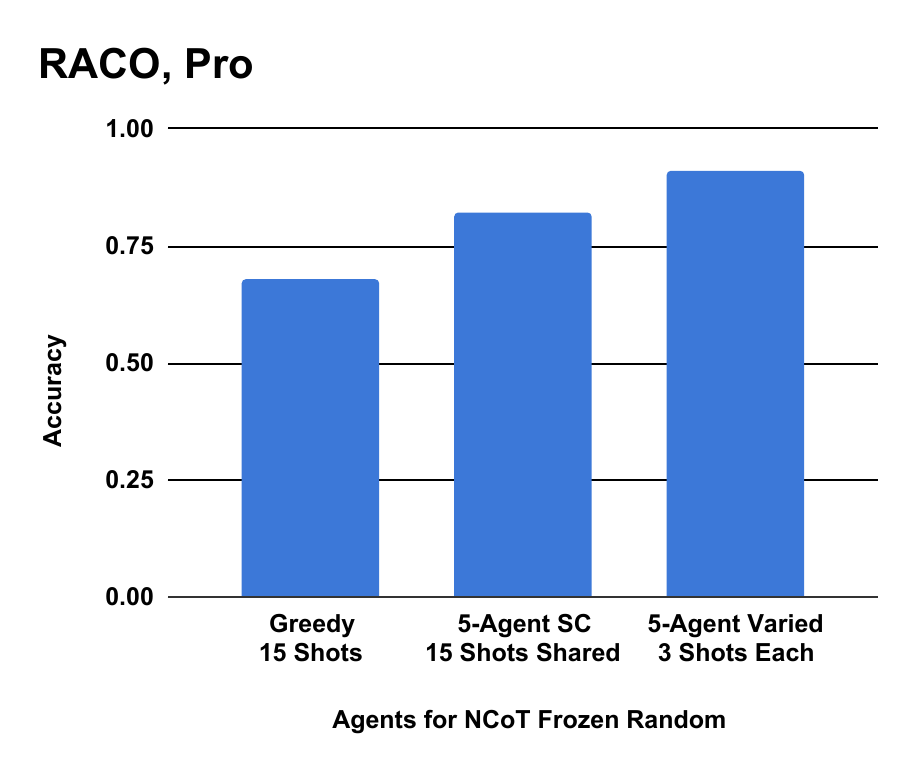}}
  \subfigure{\includegraphics[width=0.49\linewidth]{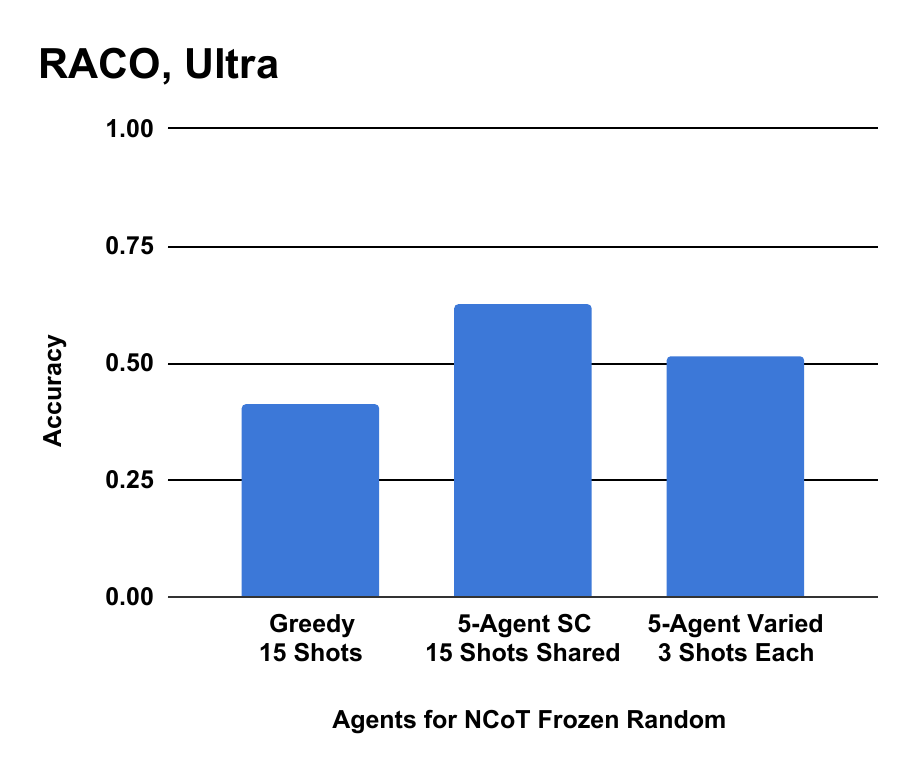}}
  \subfigure{\includegraphics[width=0.49\linewidth]{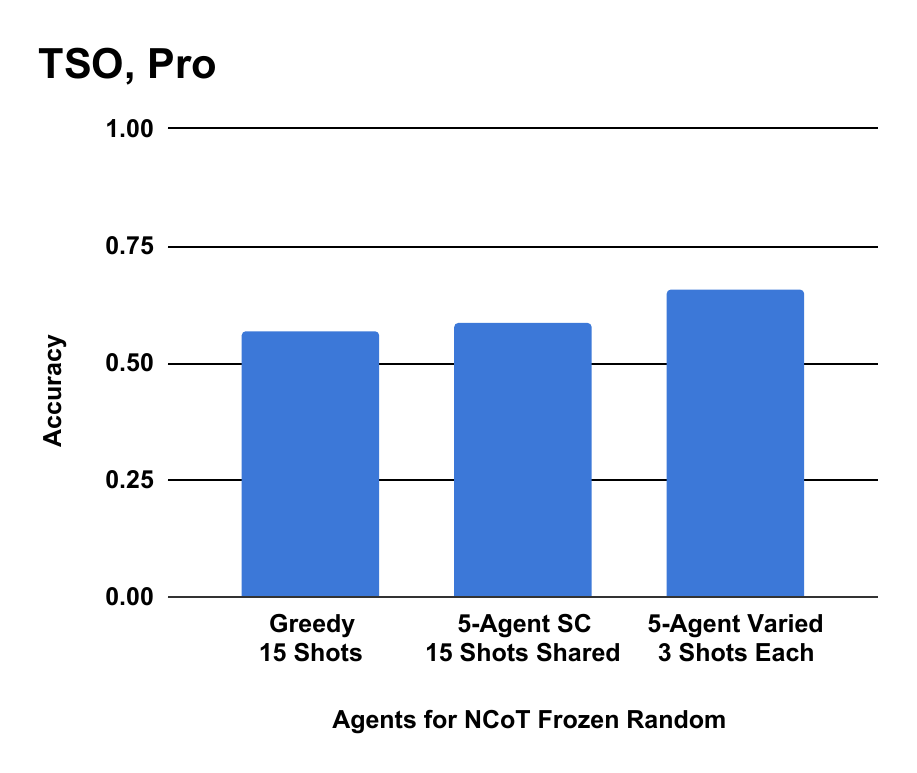}}
  \subfigure{\includegraphics[width=0.49\linewidth]{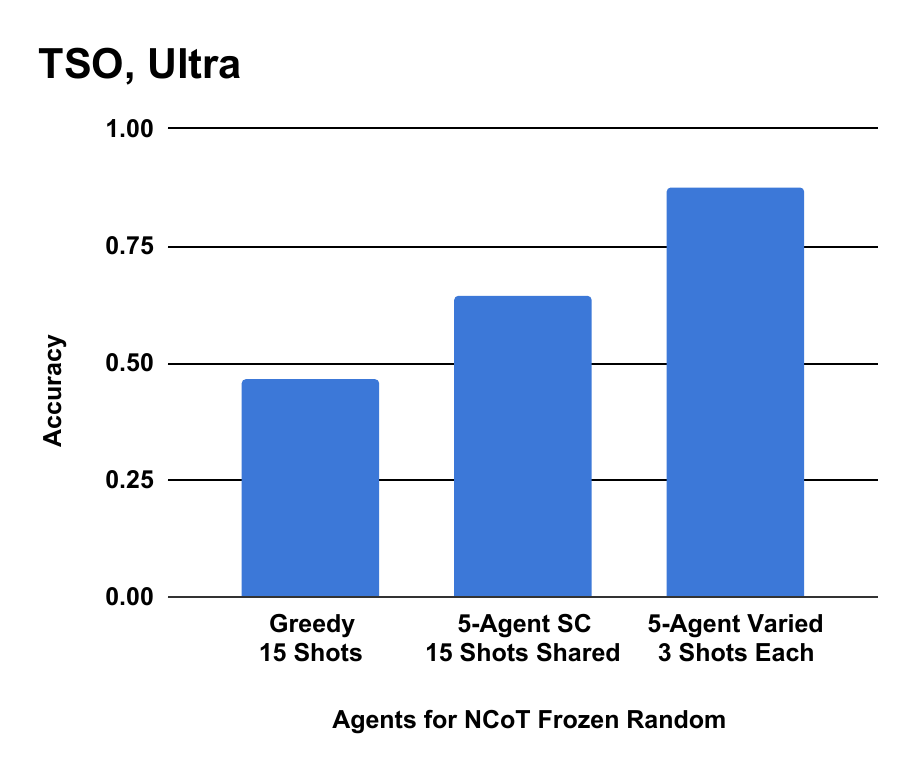}}
  \caption{More Shots vs Varied-Context Agents. Accuracy over the validation set for the two models and three tasks.}
  \label{allshots-plots}
\end{figure*}

\begin{figure*}
  \centering
  \subfigure{\includegraphics[width=0.49\linewidth]{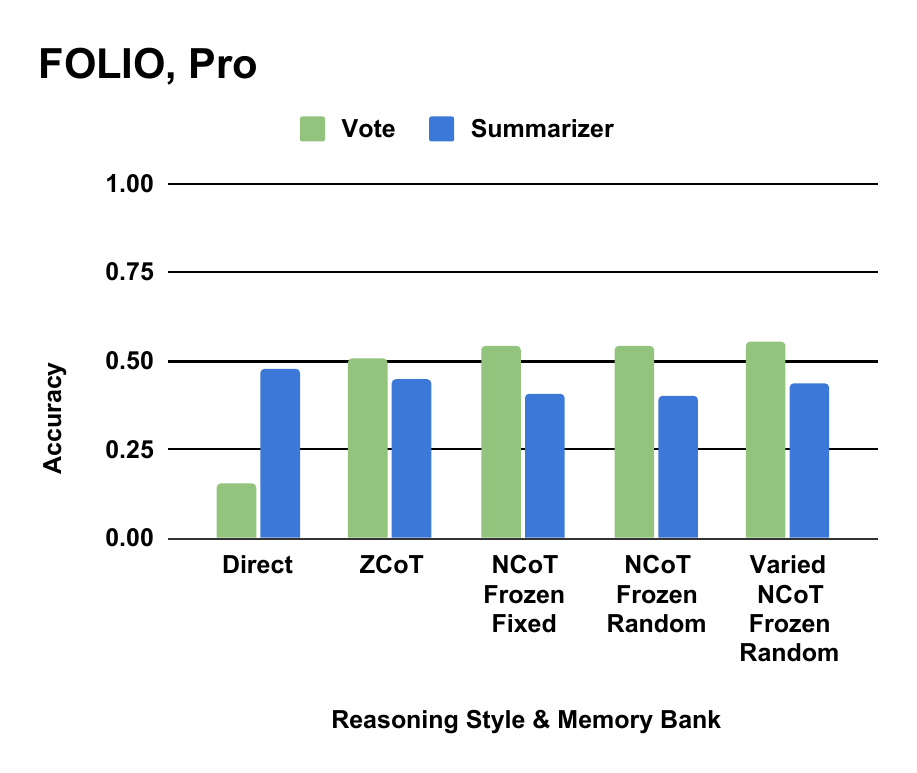}}
  \subfigure{\includegraphics[width=0.49\linewidth]{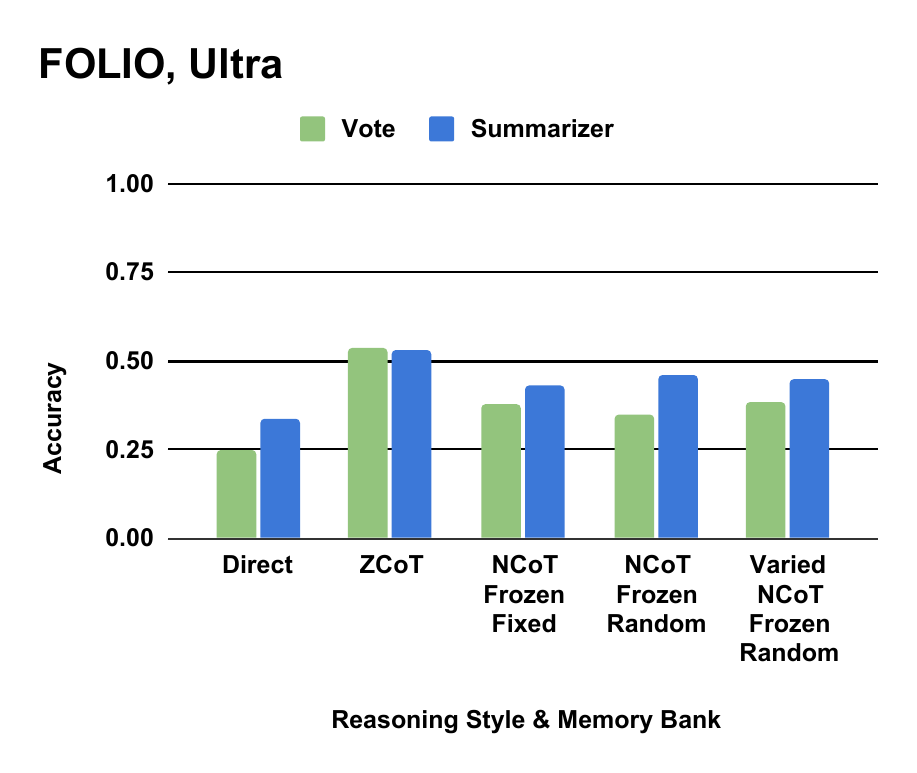}}
  \subfigure{\includegraphics[width=0.49\linewidth]{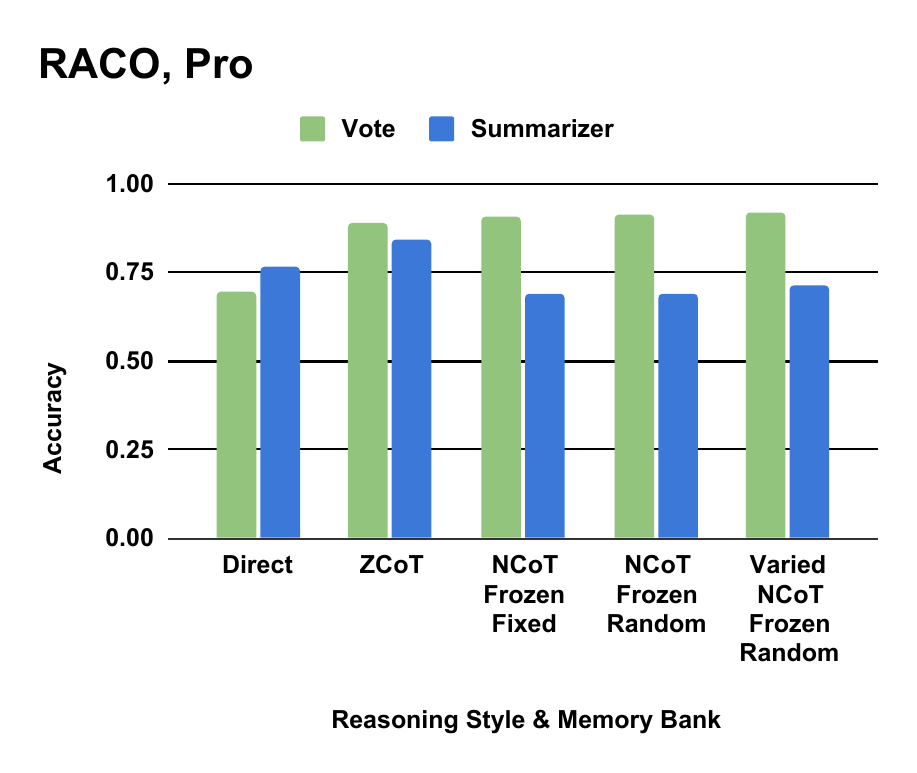}}
  \subfigure{\includegraphics[width=0.49\linewidth]{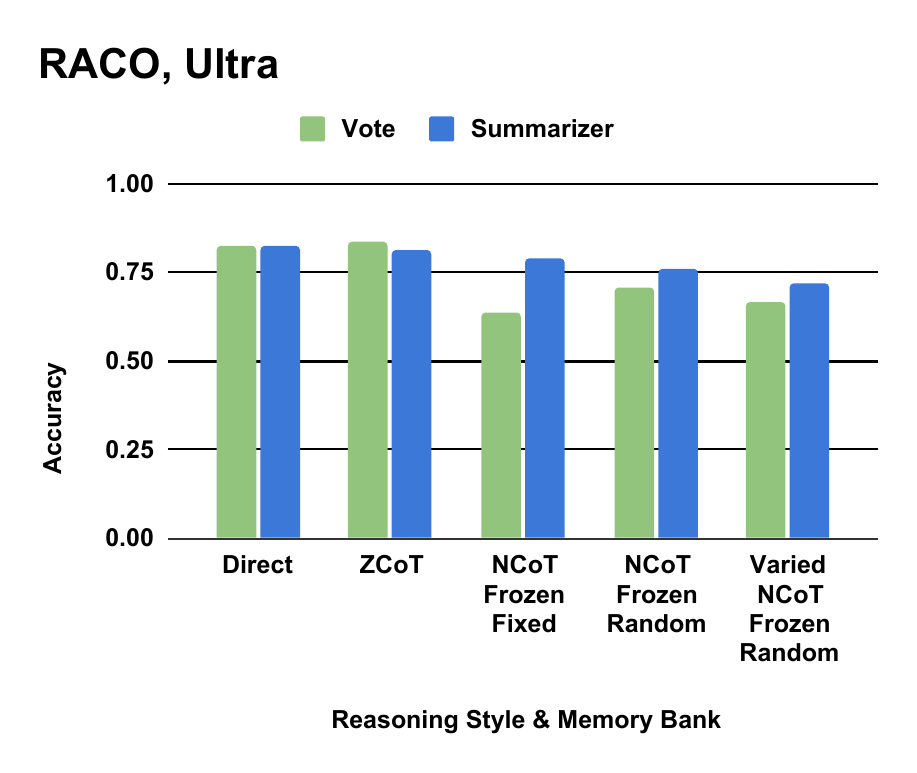}}
  \subfigure{\includegraphics[width=0.49\linewidth]{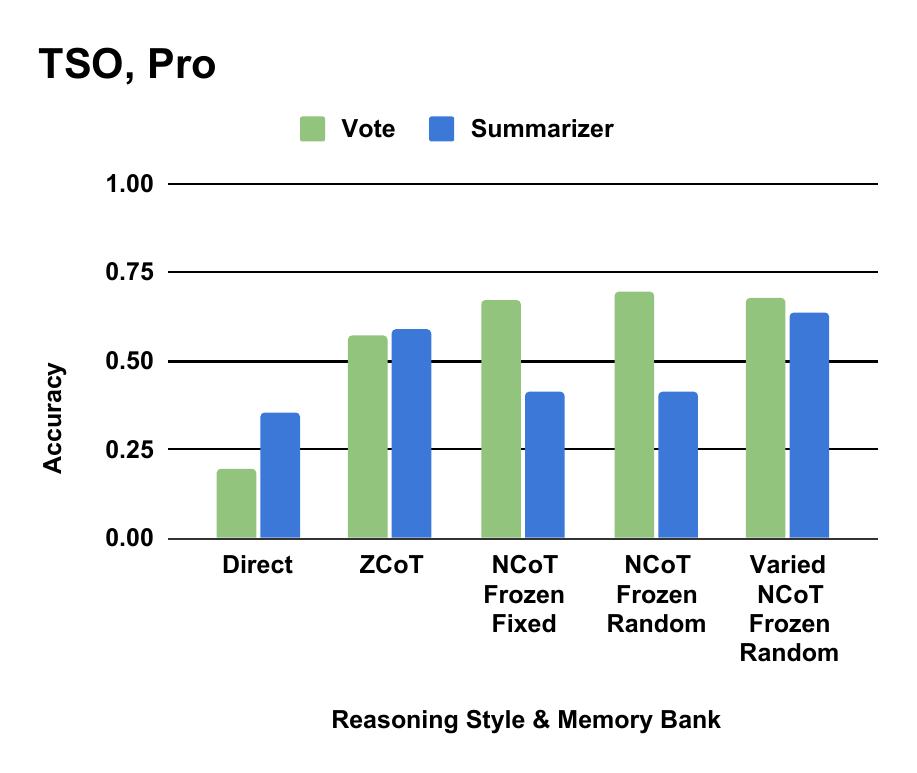}}
  \subfigure{\includegraphics[width=0.49\linewidth]{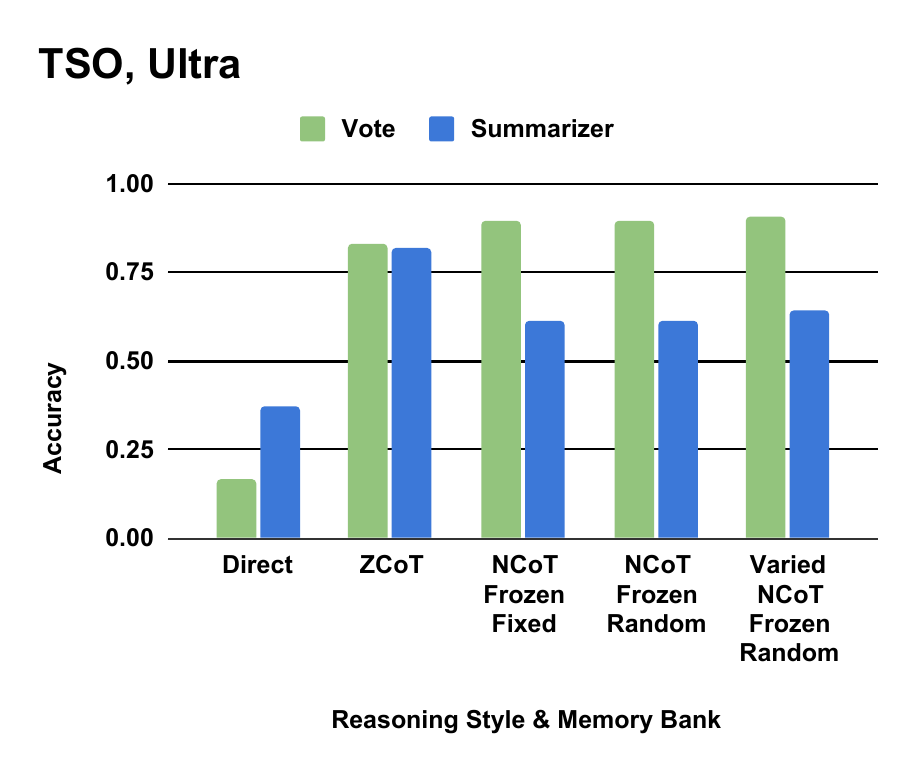}}
  \caption{Summarizer Agent vs Voting. Accuracy over the validation set for the two models and three tasks.}
  \label{summarizer-plots}
\end{figure*}

Table~\ref{main-table} corresponds to Figure~\ref{main-plots}. Figures~\ref{analogical-plots}, \ref{allshots-plots}, and \ref{summarizer-plots} show the results corresponding to  Tables~\ref{analogical-table}, \ref{allshots-table}, and \ref{summarizer-table}, respectively.

\begin{table*}
  \caption{Main experiments. Percent accuracy over the validation set, with 2-sigma error bars. Highest accuracy for each task-model pair is displayed in bold. See Figure~\ref{main-plots} and the text for further descriptions of the ablated settings.}
  \label{main-table}
  \vskip 0.15in
  \centering
  \begin{tabular}{lllp{0.085\textwidth}p{0.085\textwidth}p{0.085\textwidth}p{0.085\textwidth}p{0.085\textwidth}p{0.085\textwidth}}
    \toprule                                                                                                                                                                       
                  &                &                 & \multicolumn{6}{c}{\textbf{Reasoning Style and Memory Bank}}                                                                \\
                                                     \cmidrule(r){4-9}                                                                                                             
                  &                &                 & direct       & ZCoT                  & NCoT                  & NCoT                  & NCoT                  & NCoT         \\
                  &                &                 &              &                       & frozen                & frozen                & learned               & learned      \\
    \textbf{Task} & \textbf{Model} & \textbf{Agents} &              &                       & fixed                 & random                & random                & similar      \\
    \midrule                                                                                                                                                                       
    FOLIO         & Pro            & greedy          & $21\pm\:\:0$ & $47\pm\:\:1$          & $52\pm\:\:5$          & $51\pm\:\:2$          & $12\pm\:\:6$          & $12\pm11$    \\
                  &                & SC              & $16\pm\:\:2$ & $51\pm\:\:2$          & $54\pm\:\:4$          & $\textbf{55}\pm\:\:3$ & $\:\:1\pm\:\:4$       & $32\pm\:\:7$ \\
                  &                & varied          & $-$          & $-$                   & $-$                   & $\textbf{55}\pm\:\:3$ & $\:\:1\pm\:\:1$       & $-$          \\
                  \cmidrule(r){2-9}                                                                                                                                                
                  & Ultra          & greedy          & $25\pm\:\:0$ & $49\pm\:\:1$          & $31\pm\:\:8$          & $34\pm\:\:3$          & $26\pm\:\:7$          & $34\pm\:\:5$ \\
                  &                & SC              & $25\pm\:\:1$ & $\textbf{54}\pm\:\:1$ & $38\pm\:\:4$          & $35\pm\:\:3$          & $40\pm\:\:8$          & $38\pm\:\:4$ \\
                  &                & varied          & $-$          & $-$                   & $-$                   & $38\pm\:\:3$          & $38\pm\:\:6$          & $-$          \\
    \midrule                                                                                                                                                                       
    RACO          & Pro            & greedy          & $71\pm\:\:1$ & $84\pm\:\:0$          & $84\pm\:\:9$          & $83\pm\:\:4$          & $88\pm\:\:3$          & $83\pm\:\:3$ \\
                  &                & SC              & $70\pm\:\:1$ & $89\pm\:\:2$          & $91\pm\:\:2$          & $91\pm\:\:1$          & $92\pm\:\:2$          & $74\pm10$    \\
                  &                & varied          & $-$          & $-$                   & $-$                   & $92\pm\:\:1$          & $\textbf{94}\pm\:\:2$ & $-$          \\
                  \cmidrule(r){2-9}                                                                                                                                                
                  & Ultra          & greedy          & $82\pm\:\:0$ & $80\pm\:\:1$          & $11\pm13$             & $31\pm\:\:5$          & $79\pm\:\:5$          & $44\pm\:\:4$ \\
                  &                & SC              & $83\pm\:\:1$ & $84\pm\:\:2$          & $64\pm21$             & $71\pm\:\:4$          & $70\pm24$             & $40\pm\:\:5$ \\
                  &                & varied          & $-$          & $-$                   & $-$                   & $66\pm\:\:5$          & $\textbf{88}\pm\:\:1$ & $-$          \\
    \midrule                                                                                                                                                                       
    TSO           & Pro            & greedy          & $20\pm\:\:0$ & $49\pm\:\:1$          & $61\pm20$             & $60\pm\:\:3$          & $82\pm\:\:2$          & $82\pm\:\:6$ \\
                  &                & SC              & $20\pm\:\:2$ & $57\pm\:\:2$          & $67\pm\:\:7$          & $69\pm\:\:2$          & $36\pm44$             & $40\pm27$    \\
                  &                & varied          & $-$          & $-$                   & $-$                   & $68\pm\:\:2$          & $\textbf{85}\pm\:\:2$ & $-$          \\
                  \cmidrule(r){2-9}                                                                                                                                                
                  & Ultra          & greedy          & $16\pm\:\:0$ & $78\pm\:\:1$          & $67\pm30$             & $77\pm\:\:4$          & $92\pm\:\:1$          & $52\pm\:\:8$ \\
                  &                & SC              & $17\pm\:\:2$ & $83\pm\:\:1$          & $90\pm\:\:4$          & $90\pm\:\:1$          & $54\pm97$             & $40\pm61$    \\
                  &                & varied          & $-$          & $-$                   & $-$                   & $91\pm\:\:1$          & $\textbf{93}\pm\:\:1$ & $-$          \\
    \bottomrule                                                                                                                                                                    
  \end{tabular}
  \vskip -0.1in
\end{table*}

\section{Dataset Details}
\label{task-details}

\textbf{FOLIO} \citep{han2022folio} is a first-order logic task. It presents a series of formal logical statements then asks whether the conclusion is True, False, or Unknown. The v0.0 dataset has 1004 train examples and 204 validation examples. The last word of the response is the canonical form. Available at \url{https://github.com/Yale-LILY/FOLIO/tree/main/data/v0.0}. License: Creative Commons Attribution Share Alike 4.0 International, \url{https://github.com/Yale-LILY/FOLIO/blob/main/LICENSE}.
  
\textbf{Reasoning About Colored Objects} (RACO) from BIG-bench \citep{srivastava2022beyond} is included in BIG-Bench Hard \citep{suzgun2022challenging} as a particularly challenging task. It describes a scene involving several objects of various colors, optionally makes a modification, then asks about colors, positions, or counts. The TensorFlow Datasets (TFDS) split has 1600 train examples and 400 validation examples. To put in canonical form, convert to lower case and look for the first numeric number, then spelled-out number, and finally the first word. Available at \url{https://www.tensorflow.org/datasets/community_catalog/huggingface/bigbench#reasoning_about_colored_objects} or via \url{https://huggingface.co/datasets/google/bigbench/blob/main/bigbench.py}. License: Apache License 2.0, \url{https://github.com/google/BIG-bench/blob/main/LICENSE}.

\textbf{Tracking Shuffled Objects} (TSO) from BIG-bench \citep{srivastava2022beyond} appears in BIG-Bench Hard \citep{suzgun2022challenging} as well. It assigns several people each an object, performs a series of swaps, then asks which object one of the people has at the end. The TFDS split has 3000 train examples and 750 validation examples. The canonical form attempts to isolate the object in the model's answer. Convert the response to lower case, strip punctuation, and remove several common substrings ("at the end of the [blank]", "has", "is playing", "is dancing with", "the", "ball", and "present"). Available at \url{https://www.tensorflow.org/datasets/community_catalog/huggingface/bigbench#tracking_shuffled_objects} or via \url{https://huggingface.co/datasets/google/bigbench/blob/main/bigbench.py}. License: Apache License 2.0, \url{https://github.com/google/BIG-bench/blob/main/LICENSE}.

\section{Compute Resources}
\label{compute-resources}

\citet{huang2023large} emphasize the importance of evaluating methods with comparable inference cost. Compute resources are measured in terms of LLM calls used during memory training and inference, as well as an upper bound on memory size. Let $N_t$ and $N_v$ be the number of examples in the training and validation sets, respectively. Let $K=3$ be the default number of shots for NCoT and AP-memory. Let $M=10$ be the default number of agents for multi-agent collaboration. Let $R=6$ be the number of runs. Our direct and analogical prompts use one LLM call, while our ZCoT, NCoT, and summarizer agent prompts use two. A single greedy agent does this once per example, while multi-agent collaboration repeats it $M$ times. A summarizer agent adds two more LLM calls. Memoryless setups do not use the training set and run only on the $N_v$ validation set. Frozen memory runs single-agent ZCoT on the $N_t$ training set once per model and task. It contains at most $N_t$ exemplars. Learned memory runs over the $N_t$ training set every time. For each run it stores at most $M*N_t$ exemplars for NCoT and at most $M*K*N_t$ for AP. Table~\ref{compute-table} illustrate our compute resources.

\begin{table*}
  \caption{Computational resources by type of method. Number of LLM calls while running over the training and validation sets and maximum number of exemplars stored in memory. $N_t$ = examples in training set, $N_v$ = examples in validation set, $K$ = exemplars/shots, $M$ = reasoning agents, $R$ = runs. Frozen memory training (marked with $^*$) enjoys further savings by sharing one memory bank across methods.}
  \label{compute-table}
  \vskip 0.15in
  \centering
  \begin{tabular}{llp{0.08\textwidth}p{0.08\textwidth}p{0.08\textwidth}p{0.08\textwidth}p{0.08\textwidth}p{0.08\textwidth}}
    \toprule                                                                                                            
                    &                      & \multicolumn{6}{c}{\textbf{Reasoning Style and Memory Bank}}               \\
                                           \cmidrule(r){3-8}                                                            
                    &                      & direct  & ZCoT      & NCoT          & NCoT        & AP       & AP          \\
    \textbf{Agents} & \textbf{Size}        &         &           & frozen        & learned     &          & learned     \\
    \midrule                                                                                                            
             single & training calls       & $-$     & $-$       & $2N_t^*$      & $2N_tR$     & $-$      & $N_tR$      \\
                    & max exs stored       & $-$     & $-$       & $N_t^*$       & $N_tR$      & $-$      & $KN_tR$     \\
                    & validation calls     & $N_vR$  & $2N_vR$   & $2N_vR$       & $2N_vR$     & $N_vR$   & $N_vR$      \\
                    \cmidrule(r){2-8}                                                                                   
          multiple  & training calls       & $-$     & $-$       & $2N_t^*$      & $2MN_tR$    & $-$      & $MN_tR$     \\
                    & max exs stored       & $-$     & $-$       & $N_t^*$       & $MN_tR$     & $-$      & $MKN_tR$    \\
                    & validation calls     & $MN_vR$ & $2MN_vR$  & $2MN_vR$      & $2MN_vR$    & $MN_vR$  & $MN_vR$     \\
    \bottomrule                                                                                                         
  \end{tabular}
  \vskip -0.1in
\end{table*}

\section{Error Bars}
\label{error-bars}

We run each combination of methods, task, and model six times. Accuracy is measured over the validation set for each run then averaged across runs. Treating the validation set as fixed, the sources of variability include temperature sampling, seed to select fixed exemplars, and random retrieval of exemplars from memory. Note that for frozen memory, there is one set for each task and model, while learned memory training must be rerun every time. Plots display average accuracy and tables report the average plus or minus two sample standard deviations.

\section{Prompts}
\label{prompts}

\subsection{Zero-Shot Chain-of-Thought}
\label{zcot-prompt}

Our zero-shot chain-of-thought (ZCoT) prompt follows \citet{kojima2022large}.

\begin{verbatim}
Q: [input question]
A: Let's think step by step. [LLM thoughts]
Therefore, the answer is [LLM answer]
\end{verbatim}

\subsection{Few-Shot Chain-of-Thought}
\label{ncot-prompt}

Our few-shot chain-of-thought (NCoT) repeats the ZCoT prompt for each exemplar and the input question. This makes it compatible with exemplars generated via ZCoT. It can also use any number of exemplars, including zero, for cold-starting a learned memory bank.

\begin{verbatim}
Q: [exemplar 1 question]
A: Let's think step by step. [exemplar 1 thoughts]
Therefore, the answer is [exemplar 1 answer]

Q: [exemplar 2 question]
A: Let's think step by step. [exemplar 2 thoughts]
Therefore, the answer is [exemplar 2 answer]

Q: [exemplar 3 question]
A: Let's think step by step. [exemplar 3 thoughts]
Therefore, the answer is [exemplar 3 answer]

Q: [input question]
A: Let's think step by step. [LLM thoughts]
Therefore, the answer is [LLM answer]
\end{verbatim}

\subsection{Analogical Prompting}
\label{analogical-prompt}

Our analogical prompt follows \citet[appendix D.4]{yasunaga2023large}. We extract the answer from the last \verb+\boxed{}+ contents.

\begin{verbatim}
Your task is to tackle reasoning problems. When presented with a 
problem, recall relevant problems as examples. Afterward, proceed 
to solve the initial problem.

# Initial Problem:
[input question]

# Instructions:
Make sure to include all of the following points:

## Relevant Problems:
Recall three examples of problems that are relevant to the initial 
problem. Note that your problems must be distinct from each other 
and from the initial problem. For each problem:
- After "Q: ", describe the problem
- After "A: ", explain the solution and enclose the ultimate 
answer in \boxed{}.

## Solve the Initial Problem:
Say "Let's solve the following reasoning problem." Then formulate 
your response in the following format:
Q: Copy and paste the initial problem here.
A: Explain the solution and enclose the ultimate answer in
\boxed{} here.

[LLM response]
\end{verbatim}

For AP with a memory bank, we insert exemplars just before the LLM response. Note that the exemplars have been self-generated for previous examples using the same prompt, so they usually follow the instructed format.

\begin{verbatim}
Your task is ... [AP prompt w/o memory]

[exemplar 1]

[exemplar 2]

[exemplar 3]

[LLM response]
\end{verbatim}

\subsection{Summarizer Agent}
\label{summarizer-prompt}

We present a novel prompt for the summarizer agent. It uses the general ZCoT structure but includes instructions and context surrounding the input question. Specifically, it inserts solution candidates generated by the reasoning agents and instructs the summarizer agent to aggregate them. The solution candidates may be the reasoning agents' final answers (e.g., for direct prompting when chains-of-thought are not available) or their chains-of-thought.

\begin{verbatim}
Q: We have several solution candidates for the question below. 
Please discuss and summarize these solution candidates and output 
your best answer. [input question]
Solution 1: [agent 1 thoughts or answer]
Solution 2: [agent 2 thoughts or answer]
...
Solution n: [agent n thoughts or answer]
A: Let's think step by step. [LLM thoughts]
Therefore, the answer is [LLM answer]
\end{verbatim}

\end{document}